\documentclass{article} 
\usepackage{colm2024_conference}

\usepackage{microtype}
\usepackage{hyperref}
\usepackage{url}
\usepackage{booktabs}
\definecolor{darkblue}{rgb}{0, 0, 0.5}
\hypersetup{colorlinks=true, citecolor=darkblue, linkcolor=darkblue, urlcolor=darkblue}
\usepackage{amsmath}

\usepackage[pdftex]{graphicx}
\usepackage{float}
\usepackage{subcaption}
\usepackage{natbib}

\newif\ifcomments
\commentstrue 
\usepackage{xcolor}

\title{Transformer Circuit Faithfulness Metrics Are Not Robust}

\author{Joseph Miller\thanks{Correspondence to josephmiller101@gmail.com} \\
    FAR AI
    \And
    Bilal Chughtai \\
    Independent
    \And
    William Saunders \\
    Independent
}

\colmfinalcopy 
\begin{document}

\maketitle

\begin{abstract}
Mechanistic interpretability work attempts to reverse engineer the learned algorithms present inside neural networks. One focus of this work has been to discover `circuits' -- subgraphs of the full model that explain behaviour on specific tasks. But how do we measure the performance of such circuits? Prior work has attempted to measure circuit `faithfulness' -- the degree to which the circuit replicates the performance of the full model. In this work, we survey many considerations for designing experiments that measure circuit faithfulness by ablating portions of the model's computation. Concerningly, we find existing methods are highly sensitive to seemingly insignificant changes in the ablation methodology. We conclude that existing circuit faithfulness scores reflect \textit{both} the methodological choices of researchers as well as the actual components of the circuit - the task a circuit is required to perform depends on the ablation used to test it. The ultimate goal of mechanistic interpretability work is to understand neural networks, so we emphasize the need for more clarity in the precise claims being made about circuits. We open source a library at \href{https://github.com/UFO-101/auto-circuit}{this https URL} that includes highly efficient implementations of a wide range of ablation methodologies and circuit discovery algorithms.
\end{abstract}

\section{Introduction}

Mechanistic interpretability (MI) is a form of post-hoc interpretability that attempts to reverse engineer neural networks to provide faithful low-level explanations of model behaviour \citep{olah2020zoom}. One focus of interpretability work on transformer language models is identifying `circuits' -- subgraphs of the entire model's computational graph that are primarily responsible for the model's output on some task \citep{wang2022interpretability}; where a task is specific type of problem that a language model has to solve to output correct next-token predictions (ie. sentences that require a specific algorithm to complete correctly).

A key metric used by mechanistic interpretability (MI) researchers to quantify the quality of a `circuit' for some task is it's \textit{faithfulness} -- that is, the degree to which the circuit captures the performance of the entire model \citep{zhang2024best}. In this work, we study various small and reasonable seeming variations on methodologies for measuring circuit faithfulness and find that such variations often lead to \textit{significantly different} faithfulness scores. Faithfulness is typically measured by performing a targeted, circuit-dependent \textit{ablation} to the model, and observing the effect of this on some metric of the model's output. In the context of MI, an ablation refers to a type of intervention made on the activations of a model during its forward pass with the intended purpose of `deleting' some causal pathway(s), thereby isolating the causal effect of the circuit.

In this work, we seek to answer the questions: What do circuit faithfulness metrics actually show? To what extent are they a useful test of the circuit and to what extent are they a reflection of the experimental methodology? 

We begin by reviewing the ways in which MI researchers may vary their ablation methodology (Section~\ref{ablation-in-language-models}), providing a detailed review of methods for ablating transformer circuits. Next, we test these variations on existing circuits discovered by MI researchers (Section~\ref{faithfulness}). We provide detailed case studies of the `Indirect Object Identification' circuit by \citet{wang2022interpretability}, the `Docstring' circuit by \citet{docstring} and the `Sports Players' circuit by \citet{nanda2023factfinding}. We then go on to study `optimal circuits' (Section~\ref{optimal-circuits}) in the context of automated circuit discovery \citep{conmy2023automated} -- an emerging paradigm that aims to discover circuits algorithmically, without human input. 

We conclude with recommendations for MI researchers (Section~\ref{conclusion}). We additionally release \href{https://github.com/UFO-101/auto-circuit}{AutoCircuit}, a library containing efficient implementations of the circuit-discovery and circuit-evaluation techniques used in this paper, that is significantly faster than prior implementations we tested (see Appendix~\ref{auto-circuit} for more details).

\section{Related Work}

\textbf{Circuit Analysis}. Circuit analysis is a form of post-hoc interpretability focused on understanding the full end-to-end learned algorithm responsible for some specified narrow behaviour. A circuit is a subgraph of the full computational graph of the model that (is alleged to) implement some precise behavior. Circuits have been studied in vision models \citep{cammarata2021curve, olah2020zoom} and in toy transformer models \citep{nanda2023progress, chughtai2023toy}. More recently, the circuit analysis paradigm has achieved success in interpreting transformer language models too, with a number of papers discovering circuits implementing human understandable algorithms through ablation studies \citep{wang2022interpretability, docstring, greaterthan}. To accelerate such studies, recent work has attempted to automate the process of discovering circuits \citep{conmy2023automated, Syed2023attribution, kramar2024atp}, particularly in large language models, as circuits have historically required a large amount of researcher-effort to uncover. Prior work has suggested that ideal circuits exist on the Pareto frontier of faithfulness, completeness and simplicity (description length), as the entire network is trivially optimal for the first two criteria \citep{sharkey2024sparsify}.

\textbf{Activation Patching.} \citet{zhang2024best} recommend best practices in Activation Patching (a form of ablation, defined in Section~\ref{ablation-methodology}) for measuring circuit faithfulness in a similar work to ours. They compare single layer vs. multi-layer ablation, Resample Ablation vs. Noise Ablation and logit difference vs probability metrics when Node Patching. We study a larger set of variations in ablation methodology in this work, enumerating several more choices in methodology and arguing that different optimal circuits are defined in part by different ablation methodologies, rather than prescribing a single correct approach to ablation.

\textbf{Faithful explanations in NLP}. We are interested in explaining model behavior in a way that reflects the underlying reasoning process of the model, a criteria often referred to as faithfulness. In this work we measure faithfulness by studying the \textit{fidelity} of ablated models - the similarity of the ablated output to the outputs of the full model \citep{blackboxnlp, 10.1145/3236009, agarwal2024faithfulnessvsplausibilityunreliability}. As argued by \citet{jacovi-goldberg-2020-towards}, faithfulness should be viewed as a continuum. Any interpretation is an approximation that will necessarily fail to capture some aspects of the underlying behavior.

\textbf{Mechanistic interpretability} (MI) attempts to reverse engineer trained machine learning models to produce faithful human understandable explanations of model predictions via analysis of the low level features and algorithms implemented by the network. Circuit analysis is just one important direction in this theme of work. Besides circuit analysis, MI more broadly seeks to understand the correct frame to interpret neural network computation \citep{elhage2021mathematical, bricken2023monosemanticity, cunningham2023sparse} and to understand the learned features of models \citep{li2023emergent, tigges2023linear, gurnee2024language, bills2023language}. MI has also inspired work in steering model outputs through representation engineering \citep{turner2023activation, li2024circuit, rimsky2024steering}.

\section{Measuring Faithfulness} 
\label{ablation-in-language-models}


We follow previous works \citep{wang2022interpretability, docstring, greaterthan} in defining faithfulness of circuits as the extent to which they encapsulate the full model's computation of a particular task. These works measure faithfulness by ablating the components of the computational graph that are not in the circuit and observing the change in output of the model.

However, even within this framework, there are several important further choices when designing experiments, which we review in this section and summarise in Table~\ref{tab:tuple-table}. We also provide a summary of the approaches taken by previous works in Table~\ref{tab:previous-methodologies}.

\begin{table}[t]
\renewcommand{\arraystretch}{1.2}
\resizebox{\textwidth}{!}{%
\begin{tabular}{l l l l l l l}
\toprule
\textbf{Choice}   & \textbf{Granularity} & \textbf{Component} & \textbf{Value} & \textbf{Token positions} & \textbf{Direction} & \textbf{Set} \\ \midrule
\textbf{Examples} & Heads, MLPs          & Node               & Resample/Patch & All tokens               & Ablate Clean       & Circuit      \\
                  & Q, K, V, MLPs          & Edge               & Zero           & Specific tokens          & Restore Clean      & Complement   \\
                  & Heads, MLP Neurons              & Branch             & Mean           &                          &                    &              \\
                  & Sparse features                  &                    & Noise          &                          &                    &              \\ \bottomrule
\end{tabular}%
}
\caption{The six-tuple that defines \textit{ablation methodology} for transformer circuits.}
\label{tab:tuple-table}
\end{table}

\begin{table}[t]
\resizebox{\textwidth}{!}{
\begin{tabular}{@{}lllllll@{}}
\toprule
\textbf{Work}                                                                                                         & \textbf{Granularity} & \textbf{Component}                   & \textbf{Value}     & \textbf{Token positions}                                                                       & \textbf{Direction} & \textbf{Set} \\ \midrule
\begin{tabular}[c]{@{}l@{}}\citet{vig2020investigating}\\ (Gender Bias)\end{tabular}                                                     & Heads, Neurons       & Node                                                                           & Resample (clean)   & All tokens                                                                                     & Resample Clean     & Circuit      \\
\begin{tabular}[c]{@{}l@{}}\citet{meng2022locating}\\ (ROME)\end{tabular}                & Layers               & Node                                 & Resample (clean)   & Specific tokens                                                                                & Resample Clean     & Circuit      \\
\begin{tabular}[c]{@{}l@{}}\citet{wang2022interpretability}\\ (IOI)\end{tabular}         & Heads                & \begin{tabular}[c]{@{}l@{}}Node (evaluation) /\\ Path (discovery)\end{tabular} & Mean               & Specific tokens                                                                    & Ablate Clean       & Complement   \\

\begin{tabular}[c]{@{}l@{}}\citet{conmy2023automated}\\ (ACDC)\end{tabular}             & Heads, MLPs          & Edge                                 & Resample (corrupt) & All tokens                                                                                     & Ablate Clean       & Complement   \\
\begin{tabular}[c]{@{}l@{}}\citet{docstring}\\ (Docstring)\end{tabular}     & Heads                & Node                                 & Resample (clean)   & \begin{tabular}[c]{@{}l@{}}All tokens (evaluation)/\\ Specific tokens (discovery)\end{tabular} & Resample Clean     & Circuit      \\
\begin{tabular}[c]{@{}l@{}}\citet{greaterthan}\\ (Greaterthan)\end{tabular}                               & Heads, MLPs          & Path                                                                           & Resample (corrupt) & All tokens                                                                                     & Ablate Clean       & Complement   \\
\begin{tabular}[c]{@{}l@{}}\citet{nanda2023factfinding}\\ (Sports Players)\end{tabular} & Heads, MLPs          & Path                                 & Resample (corrupt) & Specific tokens                                                                                & Ablate Clean       & Complement  \\ \bottomrule
\end{tabular}%
}
\caption{Summary of the patching methodologies used by seven previous works. Note that each methodology differs from all of the others in at least one aspect.}
\label{tab:previous-methodologies}
\end{table}

\subsection{Ablation Methodology}\label{ablation-methodology}

In the context of MI, an ablation refers to a type of intervention made on the activations of a model during its forward pass with the intended purpose of `deleting' precise causal pathways. In the language of casual inference, we denote the ablation of all activations outside a circuit \(C\) on a model \(M\) as:

\begin{equation}\label{eq1}
F(x) = M(x \mid \text{do}(a = \tilde{a})), \, a \notin C
\end{equation}

Where \(x\) is the input to the model, \(a\) is an internal activation of the model and \(\tilde{a}\) is the ablated value of \(a\). The ablation methodology determines the types of activations and values that \(a\) and \(\tilde{a}\) can be (eg. whether \(a\) is a neuron node activation or an edge between attention heads).

Intuitively, deleting important subcomponents for some task should damage task performance, and conversely deleting unimportant sub-components should preserve task performance. As such, ablations have arisen as a commonly used tool for localizing model behaviour to specific internal model components. Ablations may be used both to \textit{find} and \textit{evaluate} mechanistic explanations of model behavior.

The concept of ablation overlaps with a related technique, \textit{activation patching}, in which activations are modified during a model's forward pass to some cached values from a different input. `Corrupted' inputs are inputs which are similar to the `clean' distribution being studied, but which have crucial differences that drastically change the output. For example, a typical `corrupt' prompt could retain the structure of a `clean' prompt, while switching a proper noun, such that the correct next token prediction is changed. \textbf{In this work we consider activation patching to be a specific type of ablation, and use the term Resample Ablation interchangeably.} But we note that in general, `patching' means editing activations to some other value, instead of `deleting' them, as ablation typically connotes.

In the remainder of this section, we review the range of ablation techniques that exist in the literature, specifically as they relate to evaluating circuits. There exist several important experimental design choices when evaluating transformer circuits via ablations. These are (1) the \textbf{granularity} of the computational graph used to represent the model, (2) what type of \textbf{component} in the graph is ablated, (3) what type of \textbf{activation value} is used to ablate the component, (4) which \textbf{token positions} are ablated, (5) the ablation \textbf{direction} (whether the ablation destroys or restores the signal) and (6) the \textbf{set} of components ablated (the circuit or the complement of the circuit). A circuit-based ablation methodology can therefore be specified as a six-tuple, and prior work has used many different combinations (Table~\ref{tab:previous-methodologies}). In this paper we argue that existing evaluations of circuits are sensitive to each of these variables.


\subsubsection{Circuit Granularity}

In this work we study circuits specified at the level of attention heads and MLPs\footnote{See \citet{thickstun2024transformer} for a brief overview of the transformer architecture.}. We also separate the input of each attention head into the Q, K and V inputs, but we omit this from our diagrams for visual simplicity. This is the most common granularity for mechanistic circuit analysis \citep{conmy2023automated, wang2022interpretability, docstring, greaterthan, nanda2023factfinding}), but previous works have also studied circuits specified at the level of layers \citep{meng2022locating}, neurons \citep{vig2020investigating}, subspaces \citep{Geiger2023FindingAB} and sparse "features" \citep{marks-etal-2024-feature}.

\subsubsection{Ablation Component Type (and Associated Model Views)}

\begin{figure}[t]
    \centering
    \begin{subfigure}[b]{0.3\textwidth}
        \centering
        \includegraphics[width=\textwidth]{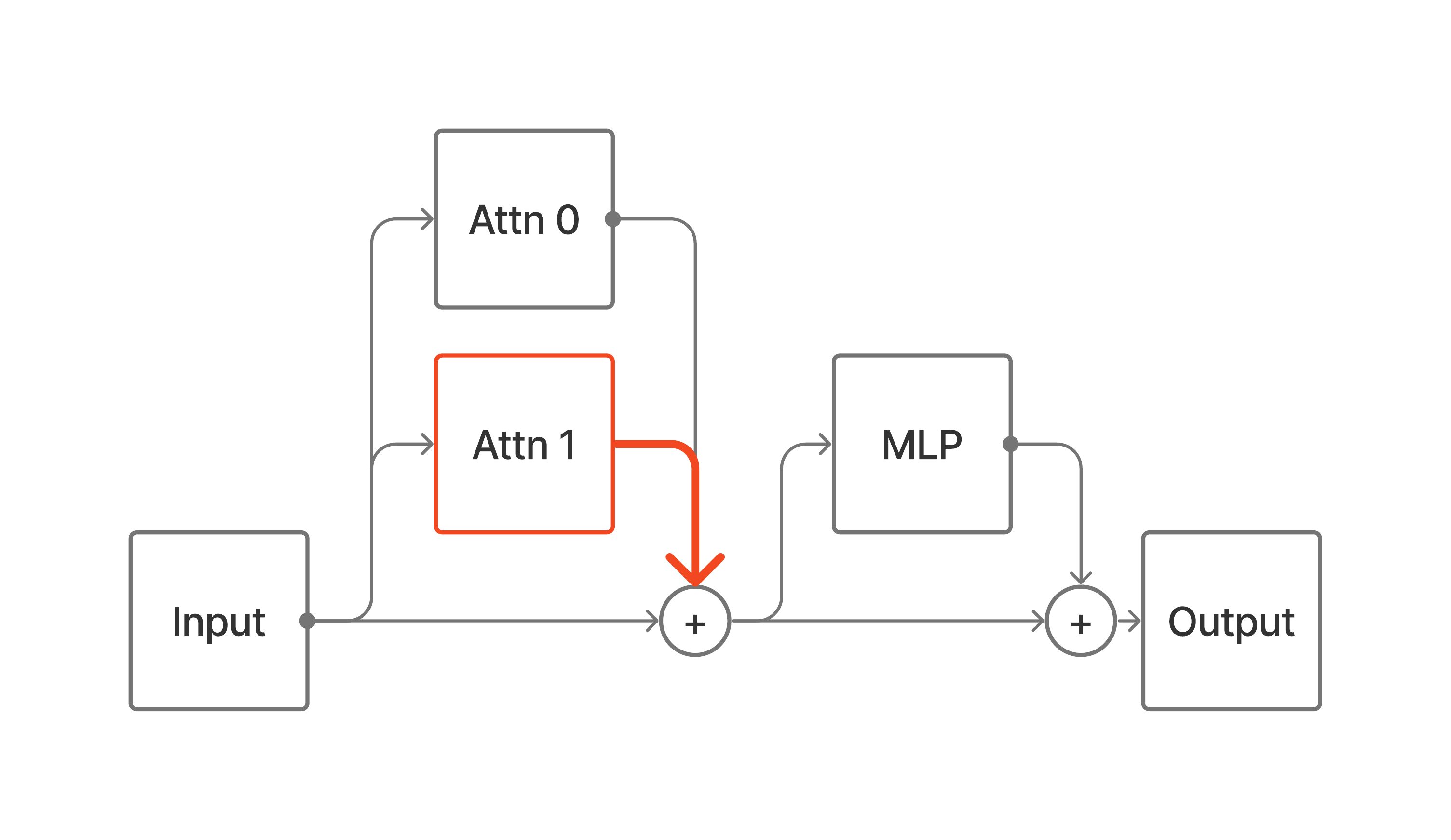}
        \caption{Node Patching (often called Activation Patching) replaces the output of some component to the residual stream in the unfactorized transformer.}
        \label{fig:node-patching}
    \end{subfigure}
    \hfill
    \begin{subfigure}[b]{0.3\textwidth}
        \centering
        \includegraphics[width=\textwidth]{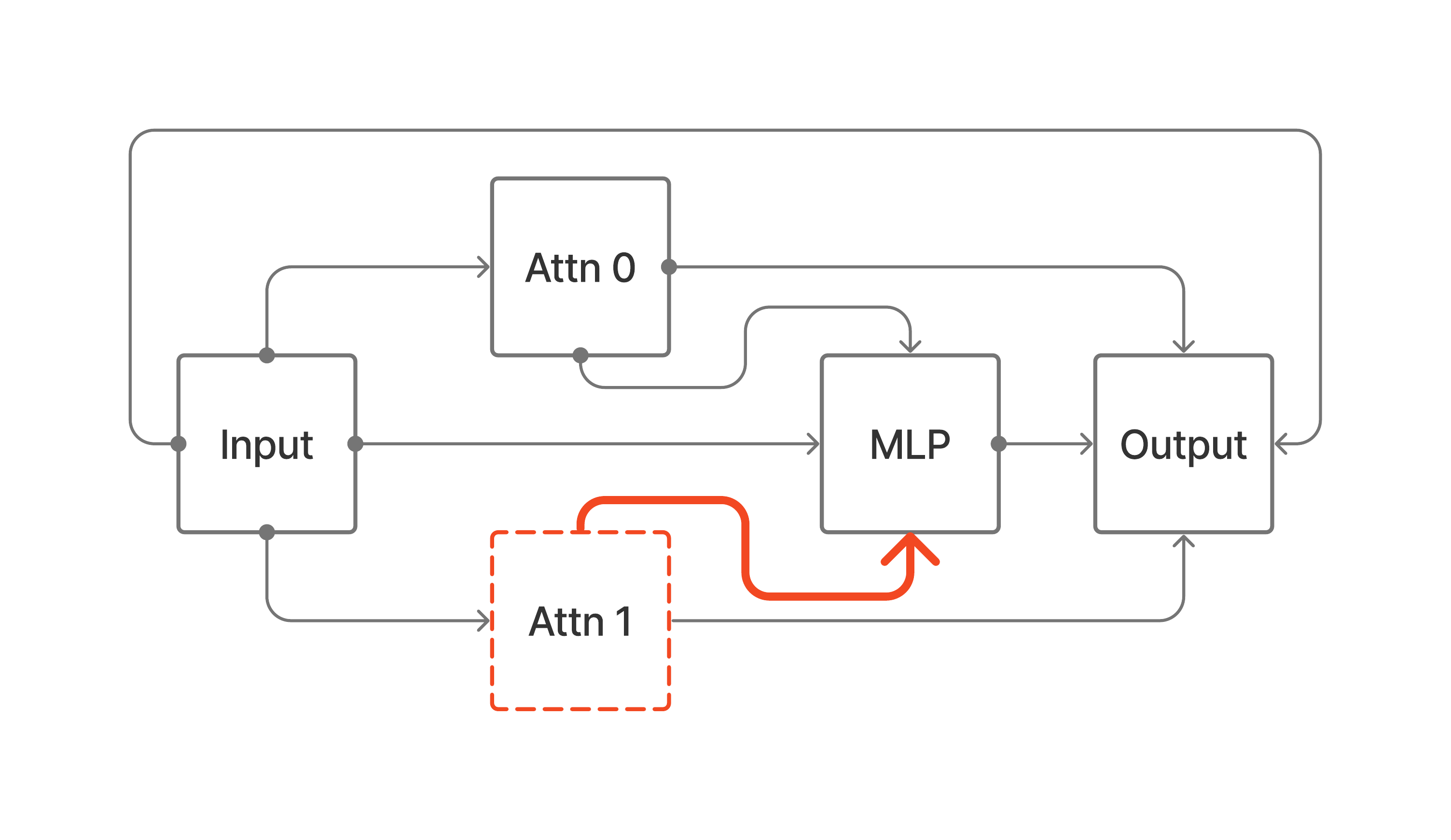}
        \caption{Edge Patching replaces the activations of a single edge in the factorized view of a transformer.}
        \label{fig:edge-patching}
    \end{subfigure}
    \hfill
    \begin{subfigure}[b]{0.3\textwidth}
        \centering
        \includegraphics[width=\textwidth]{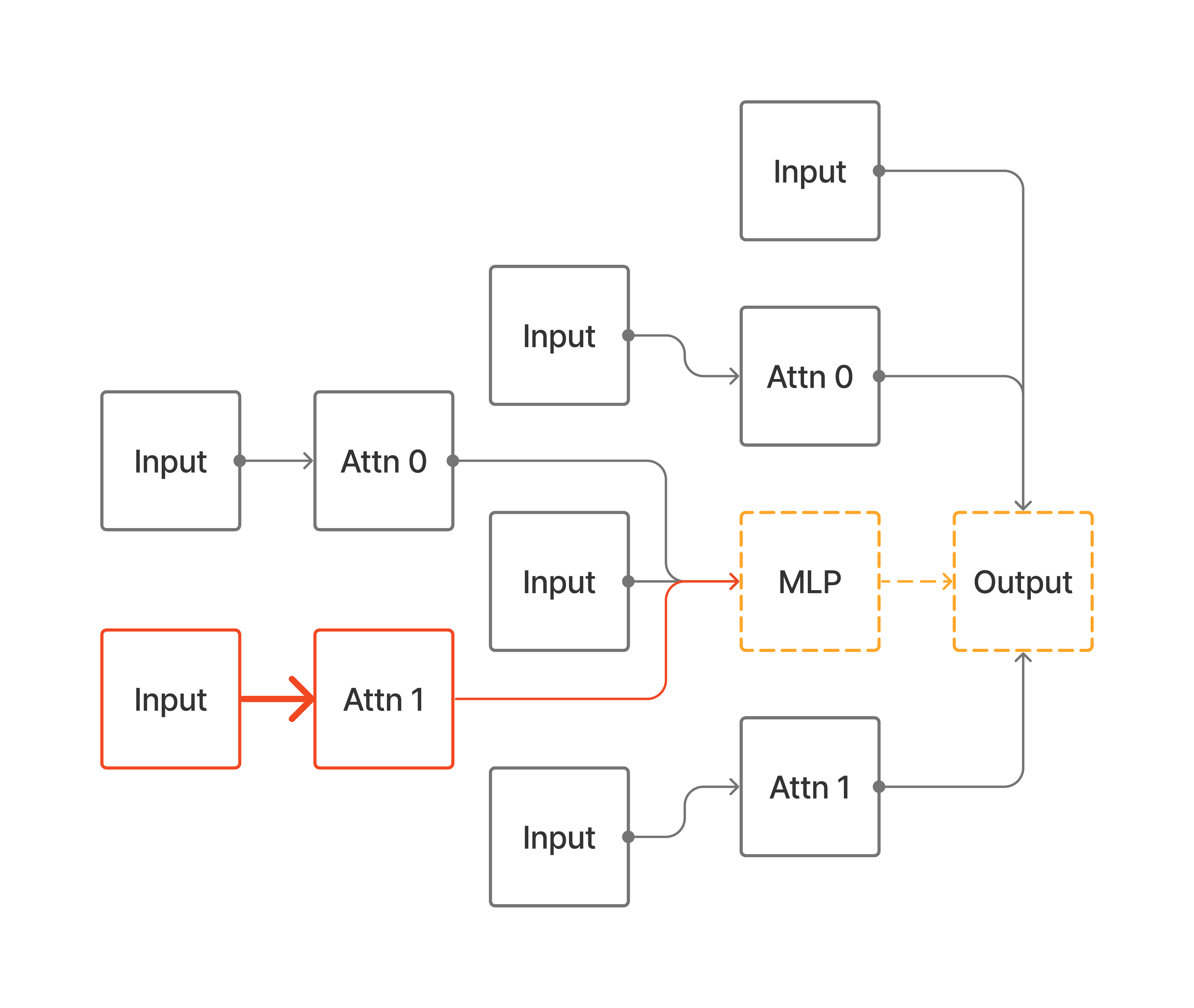}
        \caption{The `treeified' formulation of a transformer separates every path from input to output.}
        \label{fig:branch-patching}
    \end{subfigure}
    \caption{The factorized and `treeified' formulations of transformers suggest more specific ablations than ablating whole nodes.}
    \label{fig:node-edge-patching-comparison}
\end{figure}

Transformers can be described as computational graphs in several different, equivalent ways. We can choose to write the graph as a residual network (Figure~\ref{fig:residual-transformer}) or a `factorized' network in which all nodes are connected via an edge to \textit{all} prior nodes (Figure~\ref{fig:factorized-transformer}) \citep{elhage2021mathematical}. Or we can write down a `treeified' network that separates all paths from input to output (Figure~\ref{fig:treeified-transformer}). All formulations are equivalent but the `factorized' view allows us to isolate interactions between individual components and the `treeified' view allows us to isolate chains of interactions from input to output.

The component type defines the type of intervention made: we detail three possibilities, with increasing granularity. The more granular approaches are generally more difficult to implement and more computationally expensive.

\textbf{(1) Nodes}. We may intervene on a node (in the standard, residual view) during the forward pass, replacing its activation with some other value (Figure~\ref{fig:node-patching}). This is the least specific form of ablation. Since all downstream nodes `see' the change there are a large number of causal pathways affected by the ablation, which may result in unintended side-effects. This type of ablation is also known as (vanilla) activation patching \citep{vig2020investigating} when we ablate with a cached activation from another input.

\textbf{(2) Edges}. Using the factorized view of a transformer, we may intervene on an edge between two components (Figure~\ref{fig:edge-patching}). This is more specific than ablating nodes, as only the \textit{specified} destination node receives the ablated activation of the source node, so a smaller number of causal pathways are affected.

\textbf{(3) Branches}. The previous two ablations can be applied to individual nodes or edges, or to a collection of nodes and edges. Branch ablations on the other hand can only be applied to \textit{paths} from input to output (Figure~\ref{fig:branch-patching}). The causal effect of individual paths through the model is isolated by `treeifying' the factorized model. This approach was introduced by \citet{causal_scrubbing} (formalized by \citet{nix_path_patching}) and is a key component of a rigorous circuit evaluation approach known as Causal Scrubbing. However, because the number of paths in the treeified model is exponential in the number of layers of the model this approach to circuit evaluation is often intractable in practice. We omit treeified experiments in this work.

\subsubsection{Ablation Value}

When performing a causal intervention on some activation, we may choose what value we patch in. The simplest choice is to \textbf{Zero Ablate}, by replacing the activation with a vector of zeros \citep{olsson2022context, cammarata2021curve}. Prior work has noted however that the zero point is arbitrary \citep{wang2022interpretability}. The next simplest is to apply \textbf{Gaussian Noise} (GN) to the token embeddings of the clean input to obtain corrupted activations \citep{meng2022locating}. Both of these approaches can take the model significantly out of distribution \citep{zhang2024best}, producing noisy outputs \citep{wang2022interpretability}.

Two more principled approaches are \textbf{Resample Ablation} (take an activation from some other corrupted input) \citep{vig2020investigating, meng2022locating}, and \textbf{Mean Ablation} (replace with the mean activation of a node from some \textit{distribution}) \citep{wang2022interpretability}. These two ablation types have the desirable property of keeping the model closer to its usual distribution of activations. Importantly, they do not delete all information present in a component. Instead, they delete information that \textit{varies} across the distribution, while preserving information that is \textit{constant} across it, allowing us to isolate precise language tasks, while ignoring, say, generic grammar processing. When Mean Ablating, we have an additional choice in the size of the mean ablation dataset (see Section~\ref{variance-between-methodologies}). We focus on Mean and Resample Ablations in this work.

\begin{figure}[t]
    \centering
    \begin{subfigure}[b]{0.45\textwidth}
        \centering
        \includegraphics[width=\textwidth]{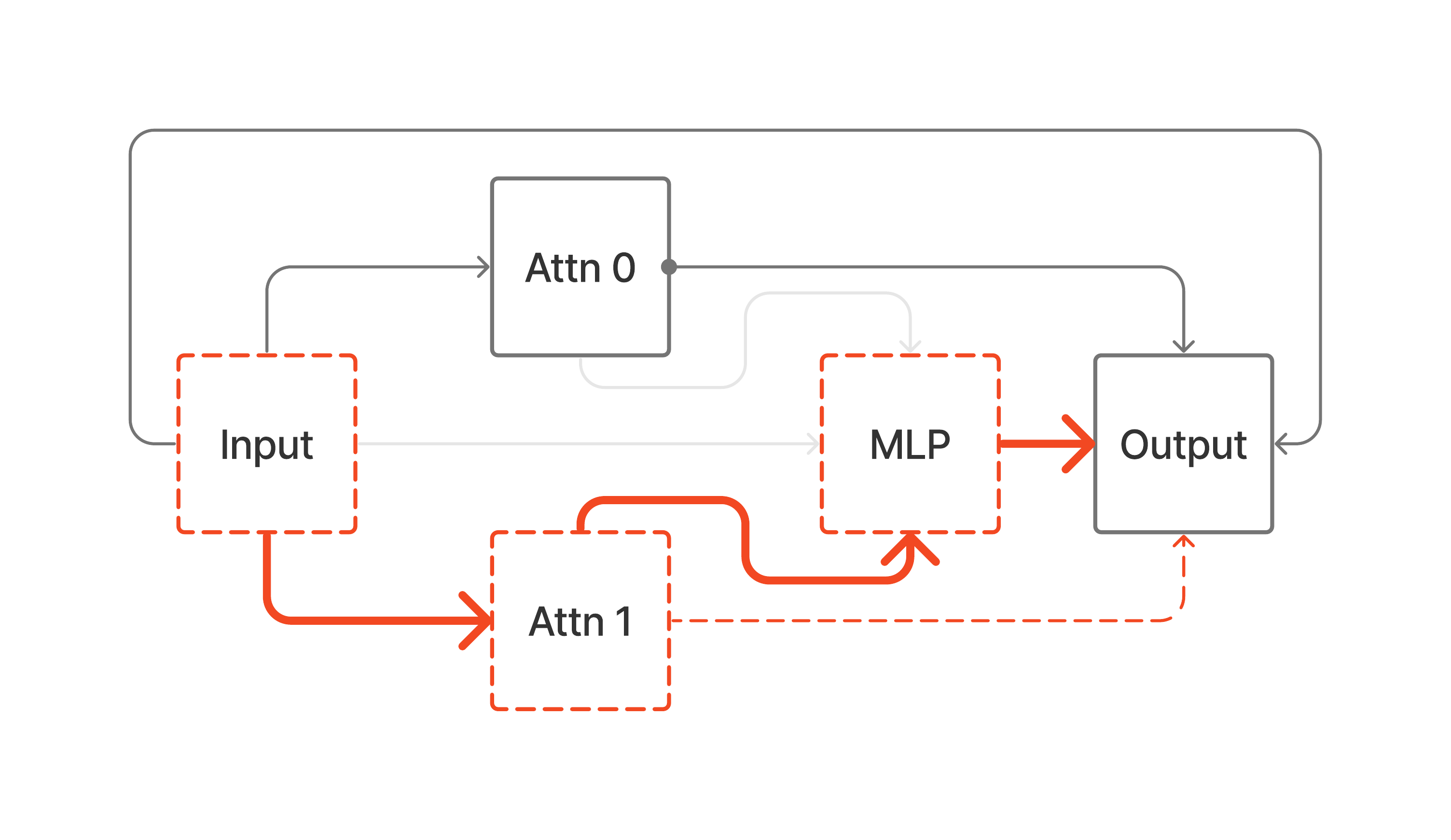}
        \caption*{(Table~\ref{tab:task-table}, Row 2) Edge Patching all the edges in a circuit with clean activations allows information from the clean input to flow along paths not included in the circuit.}
        \label{fig:edge-patching-circuit}
    \end{subfigure}
    \hfill
    \begin{subfigure}[b]{0.45\textwidth}
        \centering
        \includegraphics[width=\textwidth]{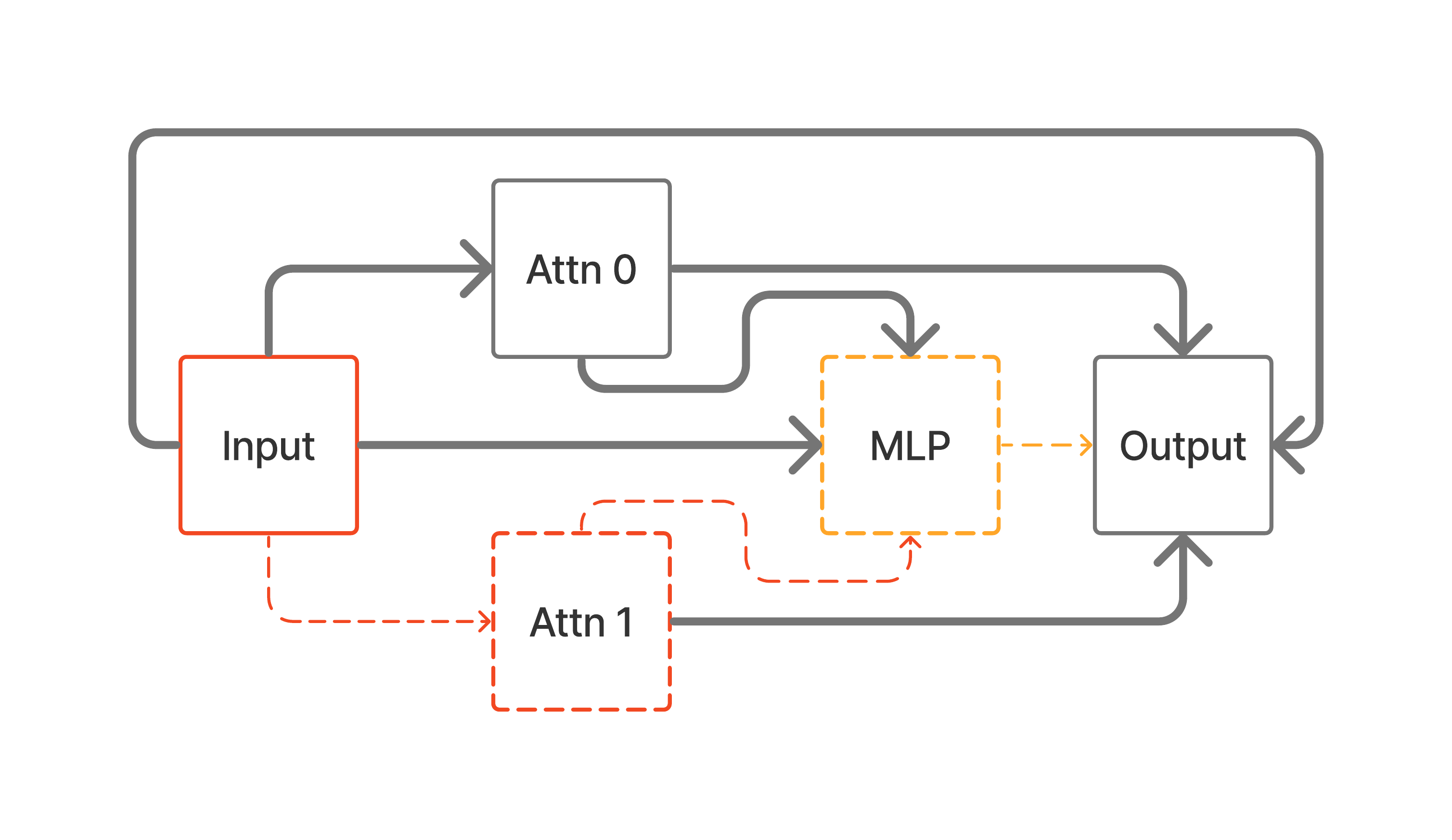}
        \caption*{(Table~\ref{tab:task-table}, Row 3) Edge Patching all the edges \textit{not} in a circuit with corrupt activations ensures that information from the clean input only flows through edges included in the circuit.}
        \label{fig:circuit-patching}
    \end{subfigure}
    \caption{Two approaches to testing a circuit that both measure faithfulness as the similarity of the output to the full model.}
    \label{fig:circuit-patching-comparison}
\end{figure}

\subsubsection{Token Positions} 
Circuits in autoregressive transformers on a narrow distribution are sometimes defined in terms of components \textit{and} token positions. When these token positions are specified, we can choose to either ablate all token positions, or only the token positions not in the specified set \citep{wang2022interpretability}. We can modify equation \eqref{eq1} to

\[
F(x) = M(x \mid \text{do}(a_i = \tilde{a}_i)), \, a_i \notin C\label{eq:1}
\]

where \(a_i\) is the activation \(a\) at token position \(i\).

\subsubsection{Ablation Direction and Testing Circuits}

Ablation typically refers to instances where we run the model on a clean input and change activations to destroy the input signal \citep{wang2022interpretability, conmy2023automated, greaterthan, nanda2023factfinding}. However, we can also run the model on a corrupt input and Resample Ablate (or Patch) in activations from the clean input \citep{meng2022locating, docstring}. Separately, when evaluating circuits, we can choose to either ablate all the components of the circuit or we can ablate all the components \textit{not} in the circuit (the complement).

The combination of these choices determines the target of our faithfulness metric:

\begin{table}[H]
\centering
\resizebox{0.7\textwidth}{!}{%
\begin{tabular}{@{}l|ll|l@{}}
\toprule
\textbf{Model Input} & \textbf{Direction} & \textbf{Set} & \textbf{Faithfulness Target} \\ \midrule
Clean                & Ablate Clean       & Circuit      & Destroy Performance          \\
Corrupt              & Restore Clean     & Circuit      & Restore Performance          \\
Clean                & Ablate Clean       & Complement   & Maintain Performance         \\
Corrupt              & Restore Clean     & Complement   & Maintain Inefficacy          \\ \bottomrule
\end{tabular}%
}
\caption{The four methodologies for directional patching for circuit evaluation.}
\end{table}

Figure~\ref{fig:circuit-patching-comparison} compares the second and third rows of the table, which both measure faithfulness as the similarity of the ablation to the full model. We note that Resample Ablating clean activations for the circuit components while passing a corrupt input \textbf{allows the signal from the clean input to flow through edges not included in the circuit}. Whereas ablating with corrupt activations on the complement of the circuit with a clean input ensures that \textbf{the signal from the input only flows through the circuit}. 


\subsection{Metric}
One further consideration in addition to the ablation methodology is the \textbf{metric} used to evaluate the effect of the ablation. We also argue that the choice of metric is important. There are many choices used in the literature, including KL Divergence \citep{conmy2023automated}, top-$k$ accuracy \citet{docstring} and task-specific benchmarks \citep{greaterthan}. In this work we will focus on the metrics used by the respective authors of the circuits that we study, but note these choices are also in general free.

\section{Faithfulness Metrics are Sensitive to Ablation Methodology}
\label{faithfulness}

In this section, we empirically demonstrate that evaluations of a given circuit's faithfulness are highly sensitive to the experimental choices outlined in Section~\ref{ablation-in-language-models} made at evaluation time. We further argue that this sensitivity is important, and may result in practitioners finding fundamentally different algorithms.

We provide a case study here on the Indirect Object Identification (IOI) circuit identified by \citet{wang2022interpretability}, as this is the most studied language model circuit in the literature \citep{conmy2023automated, makelov2023subspace, zhang2024best}, but find similar results for other known language model circuits in Appendix~\ref{app:further_faithfulness}. The IOI circuit is specified as an edge-level circuit, but \citet{wang2022interpretability} evaluate its faithfulness via a node-wise ablation methodology. We begin by testing the circuit using edge-level ablation.

\textbf{The IOI circuit}. The IOI circuit is a manually-identified subgraph of GPT-2 that is intended to perform the IOI task, which is defined by the IOI distribution. The IOI clean distribution consists of 15 sentence templates which involve two people interacting, structured such that the next word to be predicted is the indirect object \texttt{A}. Each template can be filled with names in the order \texttt{ABBA} or \texttt{BABA}, where the final \texttt{A} is the predicted token. For example: \texttt{"When John and Mary went to the store, John bought flowers for \_\_\_\_"}. The corrupt distribution (also called the ABC distribution) fills the same templates with names in the order \texttt{ABC} where \texttt{A}, \texttt{B} and \texttt{C} are three different names sampled independently of the corresponding clean prompt (we only need to specify three names because we are not defining a correct completion, unlike with \texttt{ABBA} and \texttt{BABA}). For example: \texttt{"When Gary and Nora went to the store, Naomi bought flowers for \_\_\_\_"}.

\textbf{Measuring IOI Circuit Faithfulness}. \citet{wang2022interpretability} define the \textit{metric} of circuit faithfulness to be \textbf{logit difference recovered}\footnote{\citet{wang2022interpretability} use different metrics throughout the paper. Here we are referring to the metric used to test the overall faithfulness of the circuit in Section 4 of their paper.}. The logit difference is computed between the correct answer \texttt{A} and incorrect answer (the other name in the prompt) \texttt{B} both when the full model is run as normal and when the specified nodes are ablated. Then, the percentage of the full model's logit difference which is recovered by the ablated model is calculated.

\[
\frac{F(x)_{\text{correct}} - F(x)_{\text{incorrect}}}{M(x)_{\text{correct}} - M(x)_{\text{incorrect}}} \times 100
\]

Where \(F(x)_\text{correct}\) denotes the logit of the \(\text{correct}\) answer token on \(F(x)\) (and other terms are defined similarly). A logit difference recovered of \(100\%\) means the circuit output has the same logit difference as the full model. A negative value means that the circuit outputs the corrupt logit as larger than the clean logit and a value over \(100\%\) means the circuit output has a greater logit difference than the full model. We adopt this definition of faithfulness for the remainder of this section.

\citet{wang2022interpretability} test the faithfulness of their circuit by passing in a clean input and Node Ablating the complement of the circuit. They distinguish between token positions -- that is, they ablate nodes in the circuit at all token positions except those specified by the circuit. They use a Mean Ablation, where the mean value is computed for each token position over the \texttt{ABC} distribution, using around seven examples per template.

\subsection{Variance Between Ablation Methodologies}
\label{variance-between-methodologies}

\begin{figure}[t]
    \centering
    \includegraphics[width=\textwidth]{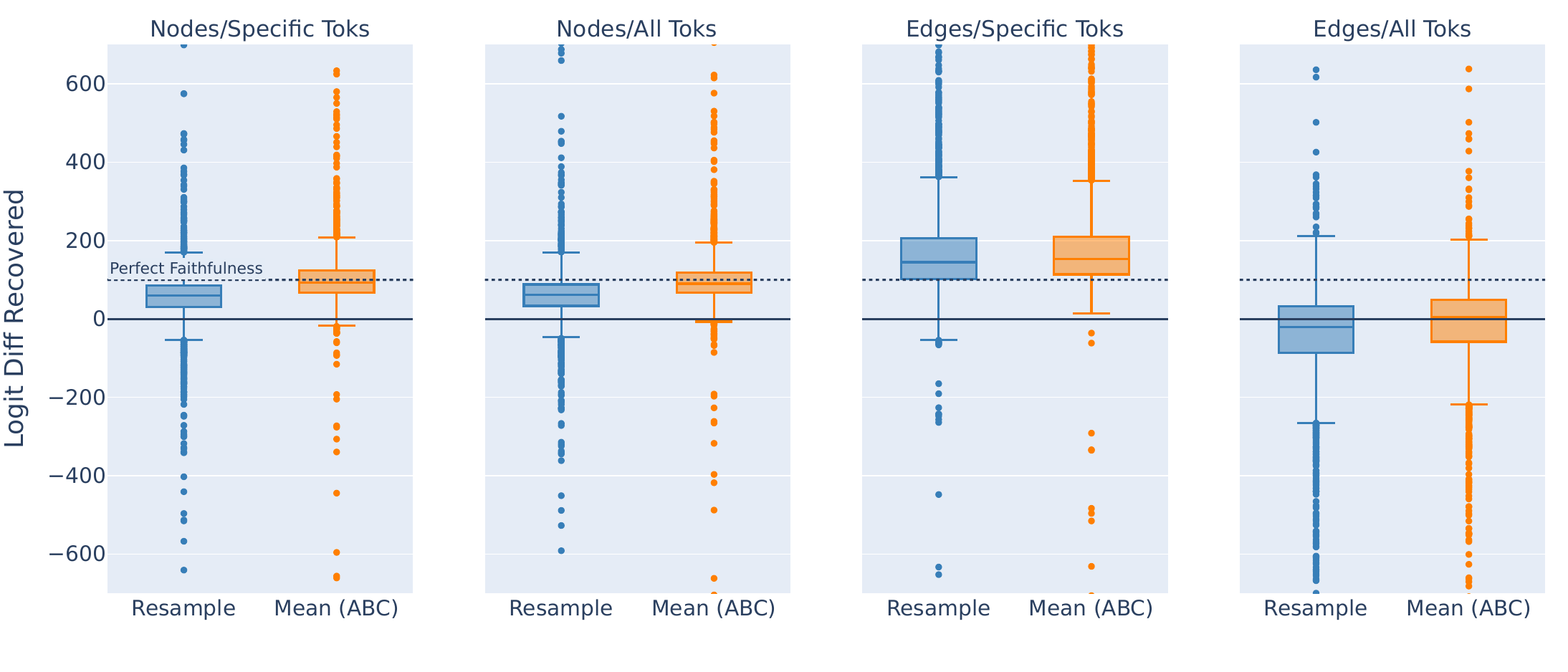}
    \caption{The IOI faithfulness metric is sensitive to (1) ablating edges/nodes, (2) the type of ablation used -- we test Resample Ablations and Mean Ablations (over a dataset of \(100\) ABC prompts, which differs from \citet{wang2022interpretability}) and (3) whether we distinguish between token positions in the circuit. The original IOI work evaluated at specific token positions with Mean Node Ablations and obtained a logit difference recovery of 87\%. Other methodologies giving faithfulness scores above 100\% or below 0\% would have given the authors significantly less confidence about the IOI circuit, and may have led them to include different edges.}
    \label{fig:ioi-edges-faithfulness}
\end{figure}

We now show circuit faithfulness is sensitive to these choices. First we compare the faithfulness metric when we change the ablation component from nodes to edges - we ablate the complement of the set of edges specified by the circuit instead of the complement of the set of nodes in the circuit. As shown in Figure \ref{fig:ioi-edges-faithfulness}, ablating at the edge level returns substantially higher percentages.

Figure~\ref{fig:ioi-edges-faithfulness} also evaluates the effect of ablation value. We rerun the above experiment using Resample Ablations from the ABC distribution, and find that this results in a systematically lower faithfulness as compared with mean ablations (statically significant on a t-test with \(p=1e-5\) for Node Ablation but not Edge Ablation). Finally, we study the effect of ablating at every token position, instead of only those specified by the circuit. This consistently results in lower faithfulness scores. It is concerning that the edge-level circuit with specific token positions has a median score well over 100\%, as this best represents the hypothesis of \citet{wang2022interpretability}.



\begin{figure}[t]
    \centering
    \includegraphics[width=\textwidth]{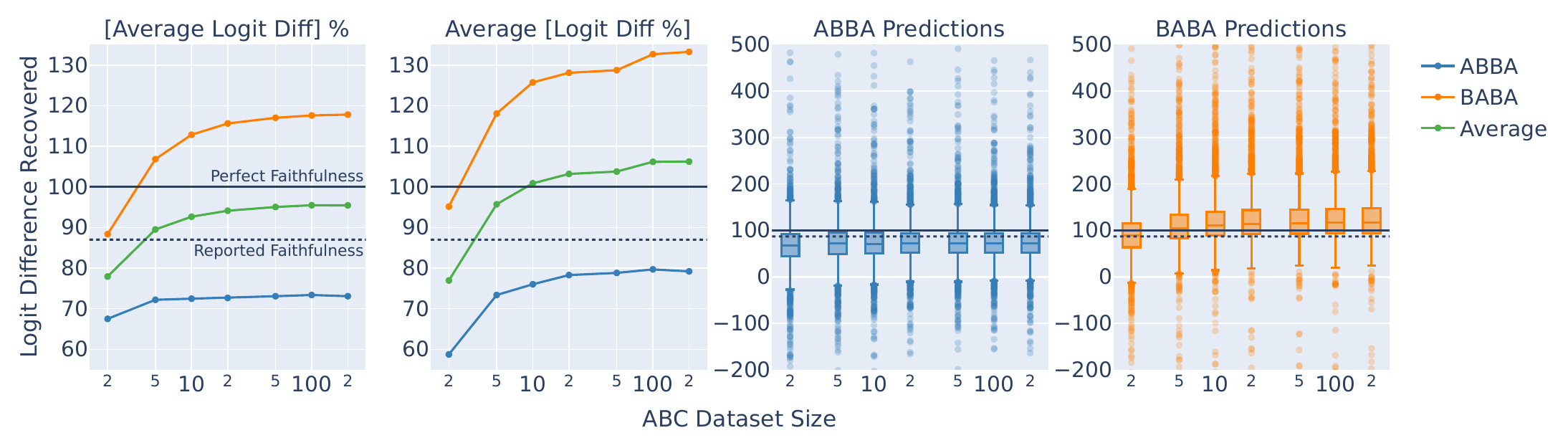}
    \caption{(Left) The IOI circuit is sensitive to the size of ABC dataset used for mean ablation. The logit difference recovered is consistently higher for prompts of the BABA format. (Left and Middle Left) The order of computing the average and percentage affects the faithfulness metric. \citet{wang2022interpretability} use [Average Logit Diff] \%, giving lower scores than Average [Logit Diff \%]. (Middle Right and Right) There is a large range of logit difference recovered, the boxplots show the interquartile range. According to this faithfulness measurement methodology, The IOI circuit implements the IOI task faithfully on average, but not for many single data points.}
    \label{fig:ioi-nodes-faithfulness}
\end{figure}

Next, we discuss sensitivity of the faithfulness metric to both the clean distribution and intricacies of the metric calculation. For these experiments, we perform node-level Mean Ablations on the complement of the circuit, split by token position, similarly to \citet{wang2022interpretability}. As shown in the left two charts of Figure~\ref{fig:ioi-nodes-faithfulness}, faithfulness is systematically greater for the prompts of form \texttt{BABA} than prompts of form \texttt{ABBA}. We also find that faithfulness monotonically increases with the size of the ABC dataset (used for computing the Mean Ablation).

Finally we note that \citet{wang2022interpretability} compute the logit difference recovered by first finding the mean logit difference for the full model and the ablated model over all prompts, and then computing the percentage (Figure~\ref{fig:ioi-nodes-faithfulness}, far left).

\[
\frac{\mathbb{E}[F(x)_{\text{correct}} - F(x)_{\text{incorrect}}]}{\mathbb{E}[M(x)_{\text{correct}} - M(x)_{\text{incorrect}}]} \times 100
\]

If instead we compute the percent difference for each prompt and then take the mean, we return substantially higher percentages (Figure~\ref{fig:ioi-nodes-faithfulness}, middle left).

\[
\mathbb{E}\left[\frac{F(x)_{\text{correct}} - F(x)_{\text{incorrect}}}{M(x)_{\text{correct}} - M(x)_{\text{incorrect}}} \times 100 \right]
\]

These are significant and important changes in evaluation. If the researchers had used a different methodology, they may have discovered a different circuit and, therefore, a different underlying algorithm. This is important since it suggests that the algorithm the circuit is required to perform depends on the ablation methodology. We expand on this point in Section~\ref{optimal-circuits}.


\subsection{Variance Between Individual Datapoints}

Even for a fixed ablation methodology and metric, there is significant variation in the measured faithfulness between individual prompts in the distribution. 

We show this for the IOI circuit in the figures above, with results for other circuits in Appendix~\ref{app:further_faithfulness}. The graphs on the right of Figure~\ref{fig:ioi-nodes-faithfulness} show a large range of faithfulness scores attained when we ablate the complement of the nodes in the IOI circuit. Note that the graphs do not show the full range of datapoints and there are several extreme outliers with a logit difference recovered in the tens of thousands of percent. The inter-quartile range (IQR) is also large, stretching up to 50\% across the dataset. This is concerning: while the circuit matches the behavior on average, it does not match it for many examples. Another property of ideal circuits describing behaviour on some task is that their faithfulness \textit{variance} should be low over the task input distribution. Otherwise, the circuit is at least partially optimized to balance out extremely high (significantly $>$100\%) and extremely low faithfulness scores ($<$0\%). This variance consideration is importantly missing from the mechanistic explanations of how GPT-2 implements the IOI task provided by \citet{wang2022interpretability}. We encourage MI researchers to evaluate task performance in both the average case and worst case.

\section{Optimal Circuits Are Defined By Prompts and Ablation Methodologies} 
\label{optimal-circuits}

We showed in the previous section that measurement details can greatly change the faithfulness score of an experiment. However, one might ask if this difference matters. In this section we discuss the consequences of such sensitivity for circuit discovery.

If a circuit is specified as a set of edges, it should be tested using edge ablations and if it is specified with token positions then it should be tested with token-specific ablation. But in other aspects there often isn't a clearly correct methodology. So how should we think about the difference in faithfulness between different methodologies? We study this question in small toy models, where we have access to the `ground truth' circuit. We conclude that the optimal circuit for some distribution cannot be defined unless we also specify the ablation methodology and metric that we are using to measure it.

Tracr models \citep{lindner2023tracr} are tiny transformers that are compiled instead of trained. Since the ground truth algorithm is both simple and known, they provide an excellent setup for testing circuit discovery algorithms. RASP programs \citep{rush2023thinkingliketransformers} are compiled into the weights of a transformer that implements the program exactly. Following \citet{conmy2023automated}, we study two Tracr models, \texttt{Reverse} and \texttt{X-Proportion}. 

The \texttt{X-Proportion} model performs the task of outputting at each token position the proportion of previous characters that are `x's. The model has two layers, with one head in each attention layer. The first attention layer and the second MLP are not used, so we need only consider the edges between the \texttt{Input}, \texttt{MLP 0}, \texttt{Attn 1.0} and \texttt{Output}. 

Conmy et al. consider the edge from \texttt{Input} to \texttt{Attn 1.0} to be part of the ground truth circuit (Figure~\ref{fig:xproportion-diagrams}). Inspecting the RASP program, we see that the only information in this edge's activation that is used by the model is the positional encoding of the tokens. However, this does not vary between different inputs, so if our ablation methodology uses Resample Ablations then this edge need not be included in the circuit, as ablating it will not change this positional information. However, if we instead use Zero Ablations, then this information will be destroyed, so the edge must be included in the circuit.

\begin{figure}[t]
    \centering
    \includegraphics[width=\textwidth]{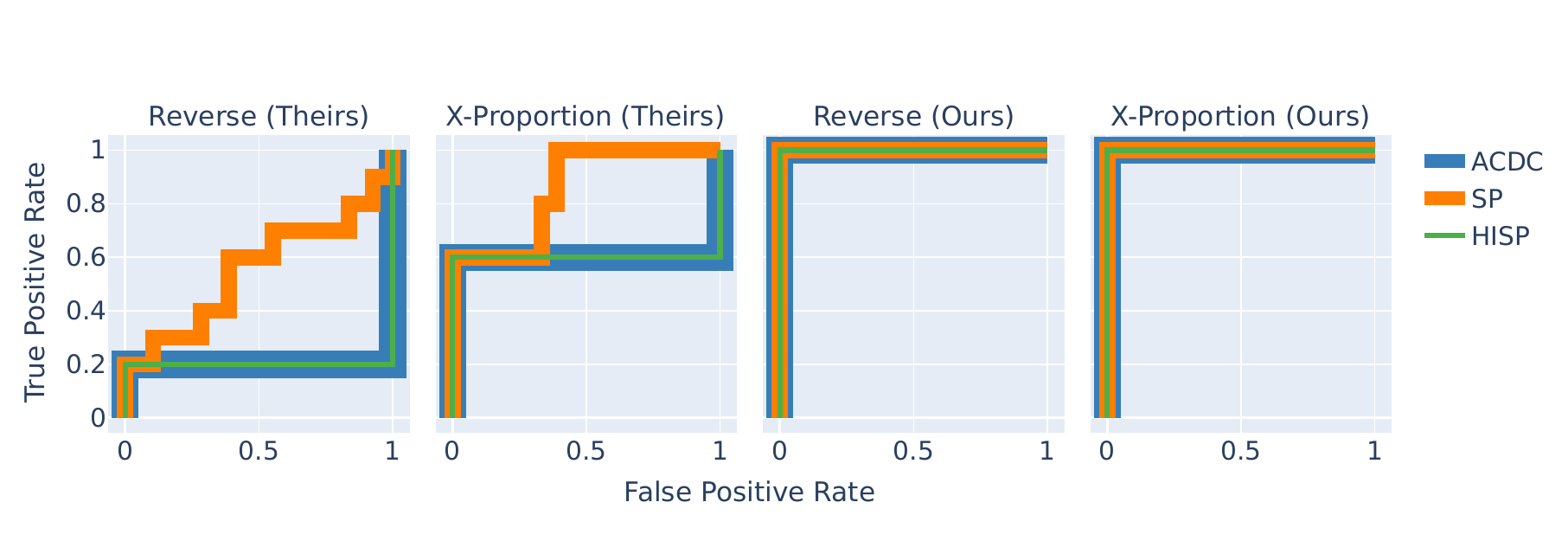}
    \caption{ROC Curves measuring the overlap between automatically discovered circuits and the two different ``ground truth'' circuits, for two Tracr tasks. When we match the ablation methodology of the ground truth with the ablation methodology of the circuit discovery algorithms, we can achieve perfect circuit recovery with all three methods.}
    \label{fig:tracr-roc}
\end{figure}

Conmy et al. test three automatic circuit discovery algorithms on this task. All three algorithms use (or approximate) Resample Ablations to discover circuits. The first method, ACDC, traverses the model in reverse topological order, ablating each edge in turn. Subnetwork Probing (SP) learns a mask parameter for each node, via gradient descent, attempting to maximize the number of nodes ablated, while minimizing the KL divergence from the original model. Lastly, Head Importance Scoring (HISP), uses a first order, gradient-based approximation of Node Ablation to assign attribution scores to each node. We test each circuit discovery method by sweeping over a range of importance thresholds to obtain an ordering of circuits of increasing size. Following Conmy et al. we then plot pessimistic receiver operating characteristic (ROC) curves (Figure~\ref{fig:tracr-roc}) and compare the area under curves.

SP and HISP, use (or approximate) Node Ablations, while ACDC uses Edge Ablations.\footnote{To convert the predictions of SP and HISP to edge-based circuits, Conmy et al. include all edges which connect two nodes of sufficient importance. With this implementation it may be impossible for SP and HISP to correctly order edges. For example, there can be two nodes which are both individually important, but where the edge connecting them is unimportant.}In our experiments we adjust the implementation of both SP and HISP to use (or approximate) Edge Ablations; SP learns mask parameters that ablate each edge and HISP assigns attribution scores for each edge by approximating Edge Patching. We provide a comparison between Edge and Node-based circuit discovery methods in Appendix~\ref{edge-vs-node-circuit-discovery}.

Conmy et al. considered the edges that would be required with Zero Ablations to be the correct circuits. Therefore, the algorithms fail to fully recover the ``ground truth". When we instead consider the edges that are required with Resample Ablations to be the correct circuit, all three algorithms perfectly recover the ``ground truth" (Figure~\ref{fig:tracr-roc}).

This case study illustrates that the optimal circuit with respect to \textit{only} a set of prompts is undefined. The ablation partly determines the task. In this case, we must decide - is determining the positional encoding part of the task? If so then the zero ablation circuit should be considered the `ground-truth', if not then the resample ablation circuit should be.

\section{Conclusion}
\label{conclusion}

In this work we show existing transformer circuit evaluations are highly sensitive to small changes in the ablation methodology and the metrics used to quantify faithfulness. We further show that the optimality of a circuit cannot be defined with respect to a set of prompts without a precise evaluation methodology

If a circuit is specified as a set of edges, it should be tested using edge ablations. And if it is specified at a chosen set of token positions it should be tested with these. But in other aspects there often isn't a clearly correct methodology. Do you want your IOI circuit to include the mechanism that decides it needs to output a name? Then use zero ablations. Or do you want to find the circuit that, given the context of outputting a name, completes the IOI task? Then use mean ablations. The task cannot be separated from the ablation methodology.

Our work has significant consequences for circuit discovery work, particularly automated circuit discovery algorithms that aim to optimize these faithfulness scores. It suggests that assessing the quality of automated methods by measuring the overlap with some `ground truth' can be misleading, if the ground truth was discovered using a different ablation methodology. 

We recommend that researchers precisely describe their experimental procedure when reporting evaluations of circuits. They should consider which task exactly they are expecting their circuit to perform.

\section{Acknowledgments}

Thanks to Arthur Conmy for his generous assistance in understanding and reproducing his work on Automatic Circuit Discovery and his insightful comments. Thanks to Adam Gleave, Lawrence Chan, Clement Neo, Alex Cloud, David Bau, Steven Bills, Sam Marks, Adrià Garriga-Alonso and our anonymous reviewers at COLM 2024 for their invaluable feedback and suggestions. Thanks to Bryce Woodworth for his help and encouragement.

\bibliography{auto-circuit.bib}
\bibliographystyle{colm2024_conference}

\appendix

\clearpage
\section{AutoCircuit Library}  \label{auto-circuit}

We release \href{https://github.com/UFO-101/auto-circuit}{AutoCircuit}, a Python library with a highly efficient implementation of Edge Patching and various circuit discovery algorithms, with support for TransformerLens models \citep{nanda2022transformerlens}. It supports Mean, Zero and Resample Ablations. See \href{https://www.lesswrong.com/posts/caZ3yR5GnzbZe2yJ3/how-to-do-patching-fast}{our blog post} for more detail on our fast implementation.

We test the performance of our implementation by running the ACDC \citep{conmy2023automated} circuit discovery algorithm, which iteratively patches every edge in the model. We compare the performance of AutoCircuit's implementation to the official ACDC implementation (which is currently the most popular library for patching large numbers of activations). We run ACDC using both libraries at a range of thresholds for a tiny 2-layer model with only 0.5 million parameters and measure the time taken to execute on a single GPU.

\begin{figure}[!ht]
    \centering
    \includegraphics[width=0.7\textwidth]{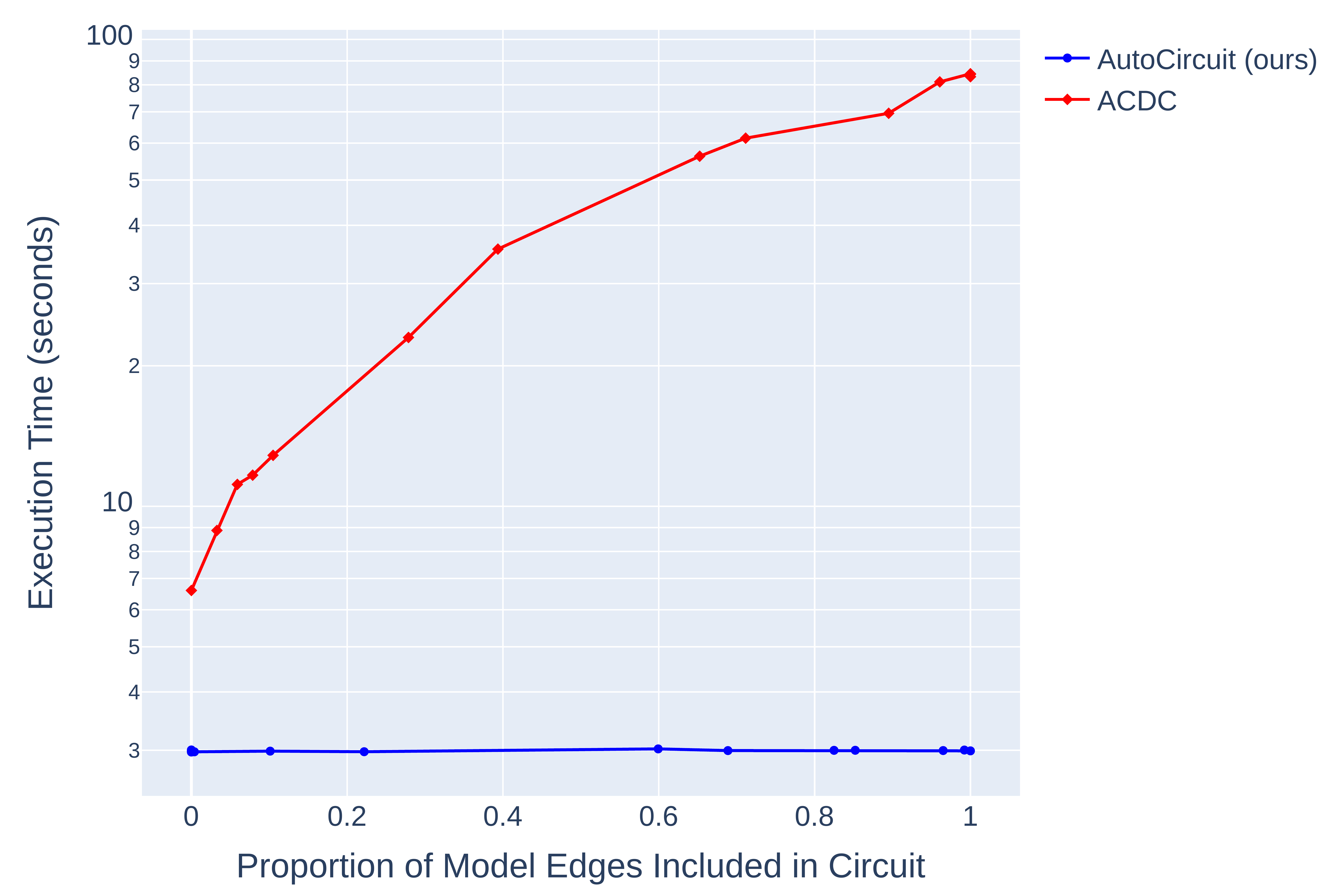}
    \caption{Time to execute the ACDC algorithm over the entire network for a small 2 layer transformer. AutoCircuit (ours) if significantly faster than the official Automatic Circuit Discovery codebase.}
    \label{fig:acdc-vs-auto-circuit}
\end{figure}

Different numbers of edges are included at different thresholds in the ACDC algorithm. Note that ACDC and AutoCircuit count the number of edges differently (AutoCircuit doesn't include 'Direct Computation' or 'Placeholder' edges) so we compare the proportion of edges included (the underlying computation graphs are equivalent). Figure~\ref{fig:acdc-vs-auto-circuit} shows that our implementation is significantly faster and the number of edges included greatly affects the performance of the official ACDC implementation, but it doesn't change the performance of our implementation.

\clearpage
\section{Further Details on Ablation Methodology} \label{further-ablation-methodology}

\begin{figure}[!ht]
    \centering
    \begin{subfigure}[b]{0.45\textwidth}
        \centering
        \includegraphics[width=\textwidth]{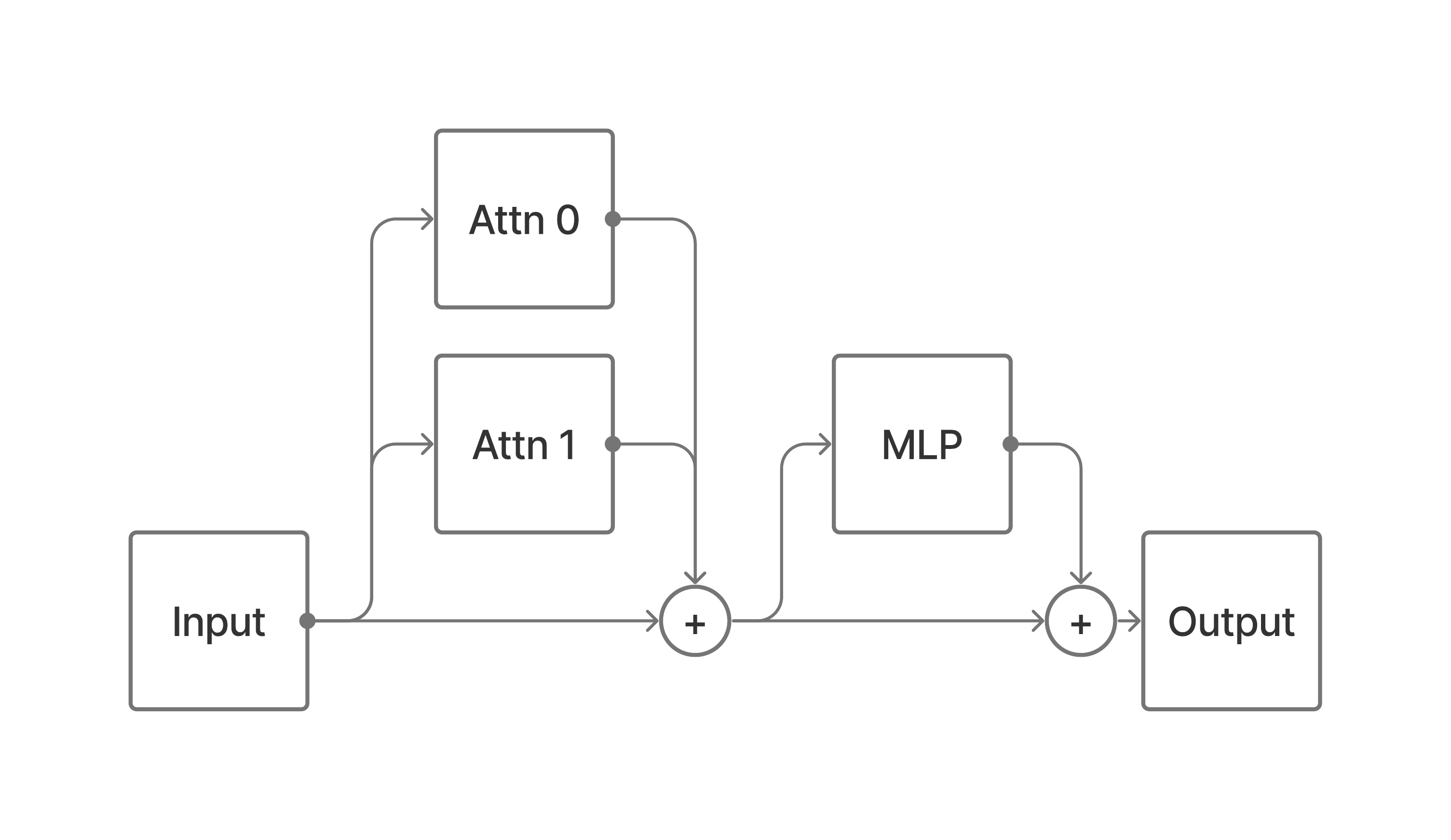}
        \caption{The canonical formulation of a transformer. Every component reads input from the residual stream backbone.}
        \label{fig:residual-transformer}
    \end{subfigure}
    \hfill
    \begin{subfigure}[b]{0.45\textwidth}
        \centering
        \includegraphics[width=\textwidth]{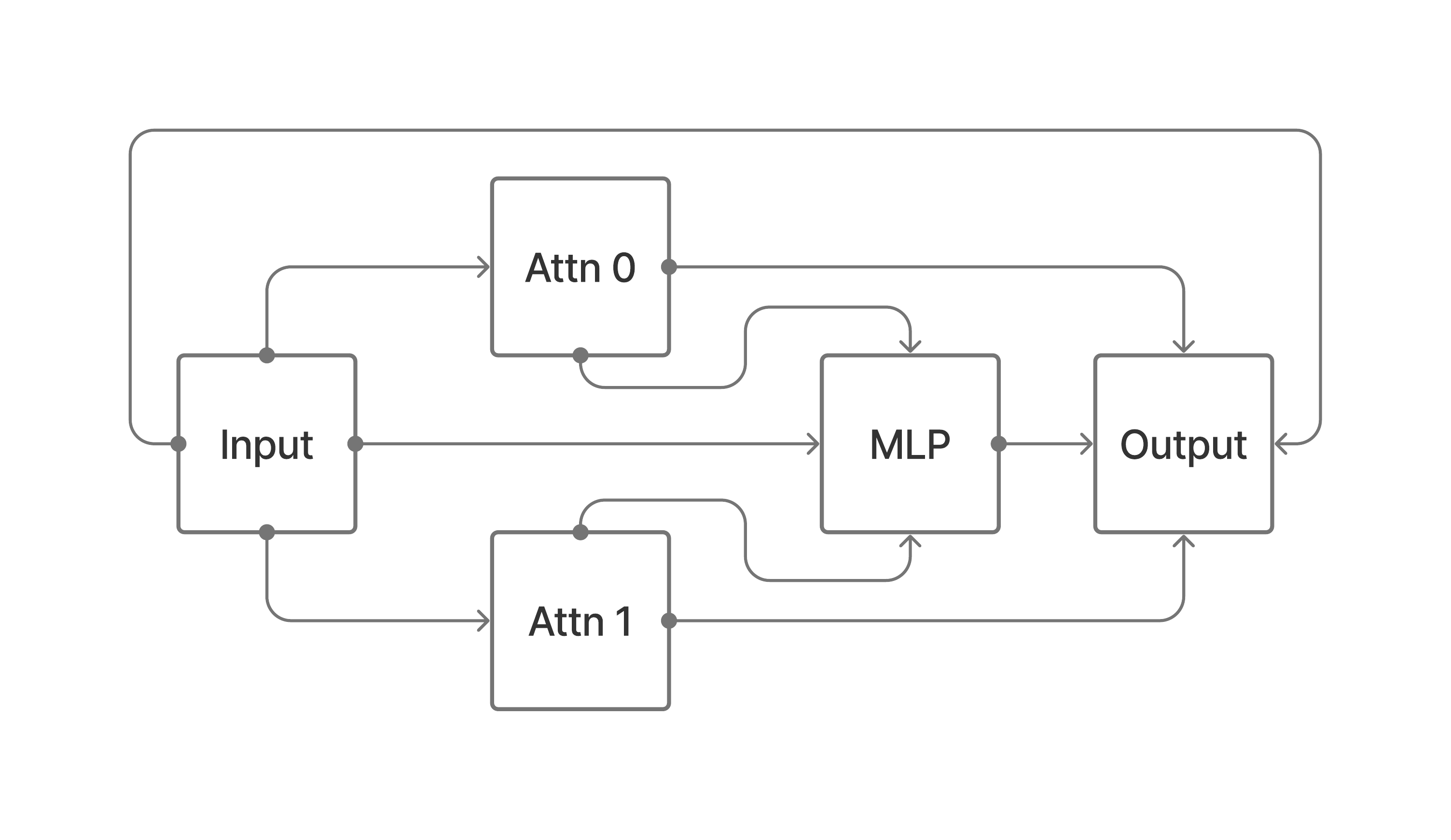}
        \caption{The ``factorized'' formulation of a transformer views every component as taking input from every previous component.}
        \label{fig:factorized-transformer}
    \end{subfigure}
    \caption{Two equivalent formulations of the transformer architecture. We illustrate only one layer, but this extends trivially to many layers.}
    \label{fig:transformer-formulation-comparison}
\end{figure}

\begin{figure}[!ht]
    \centering
    \begin{subfigure}[b]{0.45\textwidth}
        \centering
        \includegraphics[width=\textwidth]{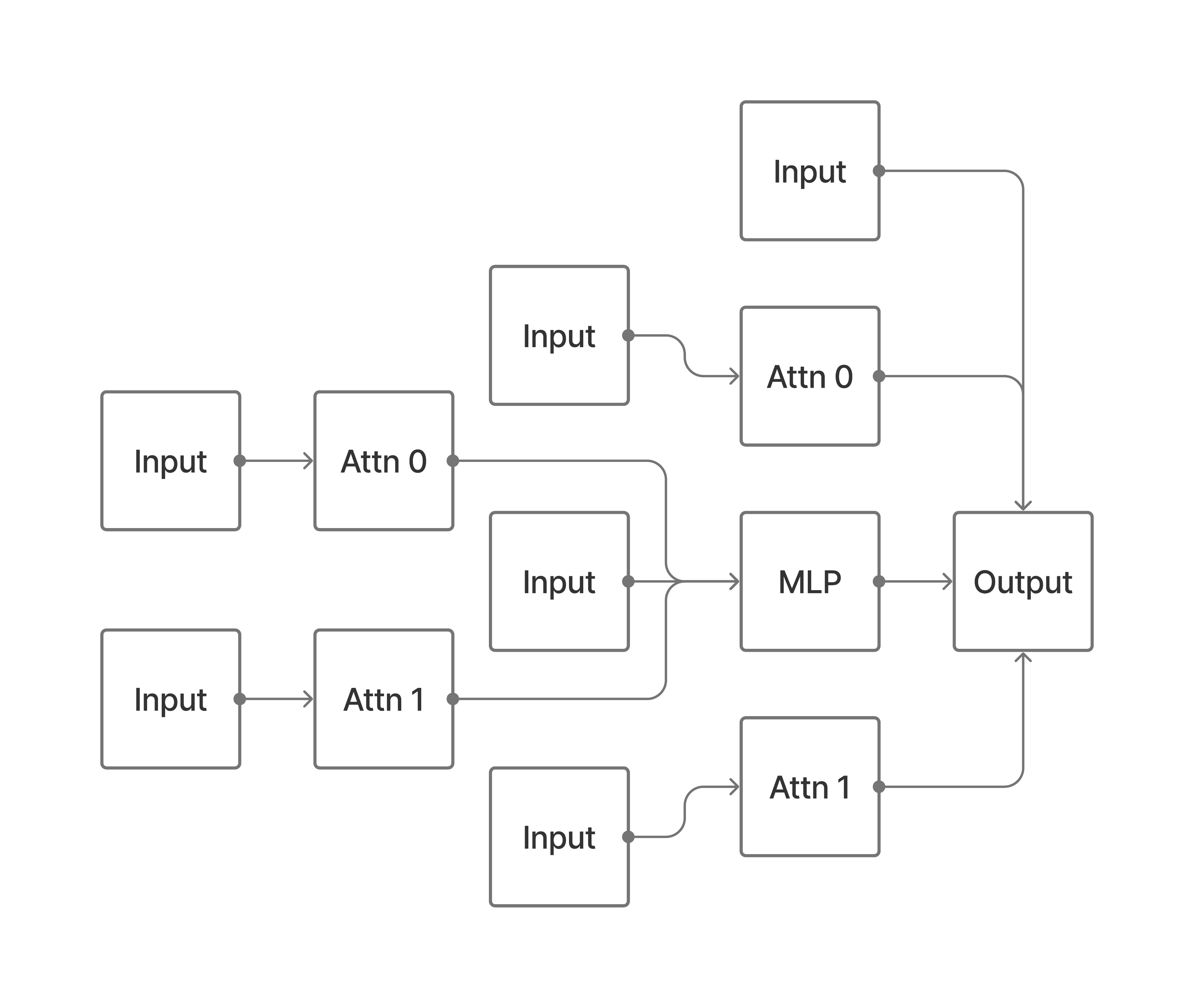}
        \caption{The ``treeified'' formulation of a transformer separates every path from input to output.}
        \label{fig:treeified-transformer}
    \end{subfigure}
    \hfill
        \begin{subfigure}[b]{0.45\textwidth}
        \centering
        \includegraphics[width=\textwidth]{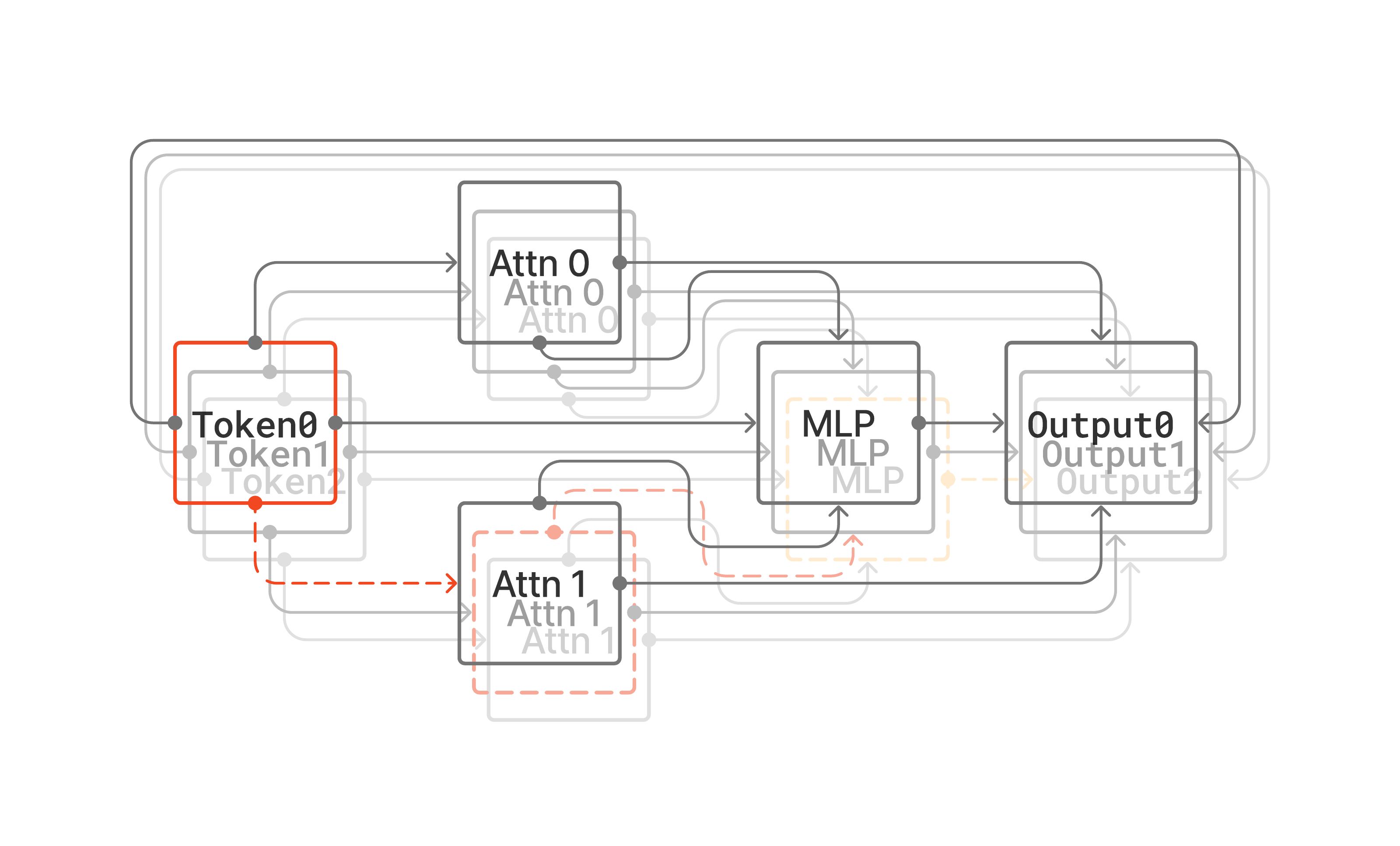}
        \caption{We can consider each token position to have a separate set of edges.}
        \label{fig:treeified-transformer-separate-tokens}
    \end{subfigure}
    \caption{(Left)``Treefied'' transformers suggest another type of ablation. (Right) Distinguishing between different token positions in Edge Patching.}
\end{figure}

\clearpage
\section{Summary of Tasks Studied}

\begin{table}[h]
\resizebox{\textwidth}{!}{
\begin{tabular}{@{}lllllll@{}}
\toprule
Name                                                                       & Model                                                              & Example Clean Prompt                                                                                                                                                                                                            & Example Corrupt Prompt                                                                                                                                                                                                              & Correct Answer                                                          & Incorrect Answer                                                                                                                                                                                & \begin{tabular}[c]{@{}l@{}}Faithfulness\\ Metric\end{tabular}             \\ \midrule
\begin{tabular}[c]{@{}l@{}}Tracr\\ X-Proportion\end{tabular}               & \begin{tabular}[c]{@{}l@{}}Tracr\\ X-Proportion\end{tabular}       & \textbf{y,x,z,x,w}                                                                                                                                                                                                              & \textbf{z,w,w,y,x}                                                                                                                                                                                                                  & \textbf{\begin{tabular}[c]{@{}l@{}}0,0.5,0.333,\\ 0.5,0.4\end{tabular}} & \textbf{0,0,0,0,0.2}                                                                                                                                                                            & \begin{tabular}[c]{@{}l@{}}Mean squared\\ error\end{tabular}              \\ \midrule
\begin{tabular}[c]{@{}l@{}}Tracr\\ Reverse\end{tabular}                    & \begin{tabular}[c]{@{}l@{}}Tracr\\ Reverse\end{tabular}            & \textbf{1,0,2,2,2}                                                                                                                                                                                                              & \textbf{1,0,0,1,2}                                                                                                                                                                                                                  & \textbf{2,2,2,0,1}                                                      & \textbf{2,1,0,0,1}                                                                                                                                                                              & KL Divergence                                                             \\ \midrule
\begin{tabular}[c]{@{}l@{}}Indirect\\ Object\\ Identification\end{tabular} & GPT-2                                                              & \begin{tabular}[c]{@{}l@{}}Then, Scott and Jeremy went\\ to the hospital. Jeremy gave a\\ snack to\end{tabular}                                                                                                                 & \begin{tabular}[c]{@{}l@{}}Then, Michael and Anderson\\ went to the hospital. Rachel\\ gave a snack to\end{tabular}                                                                                                                 & \textbf{" Scott"}                                                       & \textbf{" Jeremy"}                                                                                                                                                                              & \begin{tabular}[c]{@{}l@{}}Logit\\ Difference\\ Recovered\end{tabular}    \\ \midrule
Docstring                                                                  & \begin{tabular}[c]{@{}l@{}}4 Layer\\ Attention\\ Only\end{tabular} & \textbf{\begin{tabular}[c]{@{}l@{}}def error(self, create, option,\\ file, run, client, project):\\     """land employment camp\\ \\     :param file: protein author\\     :param run: forest degree\\     :param\end{tabular}} & \textbf{\begin{tabular}[c]{@{}l@{}}def error(self, create, option,\\ output, host, label, project):\\     """land employment camp\\ \\     :param first: protein author\\     :param text: forest degree\\     :param\end{tabular}} & \textbf{" client"}                                                      & \textbf{\begin{tabular}[c]{@{}l@{}}" size", " output", \\ " host"," label", \\ " first", " text",\\ " request", " user", \\ " file"," run", \\ " create", " option",\\ " project"\end{tabular}} & \begin{tabular}[c]{@{}l@{}}Correct\\ Prediction\\ Proportion\end{tabular} \\ \midrule
Sports Players                                                             & Pythia 2.8B                                                        & \begin{tabular}[c]{@{}l@{}}Fact: Tiger Woods plays the\\ sport of golf\textbackslash{}nFact: Phil\\ Simms plays the sport of\end{tabular}                                                                                       & \begin{tabular}[c]{@{}l@{}}Fact: Tiger Woods plays the\\ sport of golf\textbackslash{}nFact: Babe Ruth\\ plays the sport of\end{tabular}                                                                                            & \textbf{" football"}                                                    & \textbf{\begin{tabular}[c]{@{}l@{}}" basketball",\\ " baseball"\end{tabular}}                                                                                                                   & \begin{tabular}[c]{@{}l@{}}Top Sport\\ Logit\end{tabular}                 \\ \bottomrule
\end{tabular}%
}
\caption{The tasks we study, which previous works have found circuits for, and the metrics used by previous works to measure their faithfulness.}
\label{tab:task-table}
\end{table}

\clearpage
\section{Further Study of Faithfulness Metrics} \label{further-faithfulness-metrics}
\label{app:further_faithfulness}

In this section, we provide further analysis demonstrating faithfulness metrics are brittle, on two other circuits from the existing literature.

\subsection{Docstring}

\textbf{The Docstring Task}. The Docstring task \citep{docstring} is a simple task that tests a 4 layer, attention-only model's ability to complete a specific part of a standard Python docstring (see Table~\ref{tab:task-table} for an example). All prompts follow a very similar format, with the only difference being the names of the variables in the function. The corrupt distribution follows the exact same format, using a disjoint set of variable names.

\textbf{Measuring Docstring Circuit Faithfulness}. \citet{docstring} test their circuit using a similar methodology to the one which \citet{wang2022interpretability} used to test the IOI circuit. They ablate all nodes in the complement of their circuit. However, unlike \citet{wang2022interpretability} they use a Resample Ablation (also known in this context as Activation Patching), and they do not distinguish different token positions. The metric that they use for faithfulness is the percent of highest logit outputs that are the correct answer over some set of prompts.

\begin{figure}[!ht]
    \centering
    \begin{subfigure}[b]{0.45\textwidth}
        \centering
        \includegraphics[width=\textwidth]{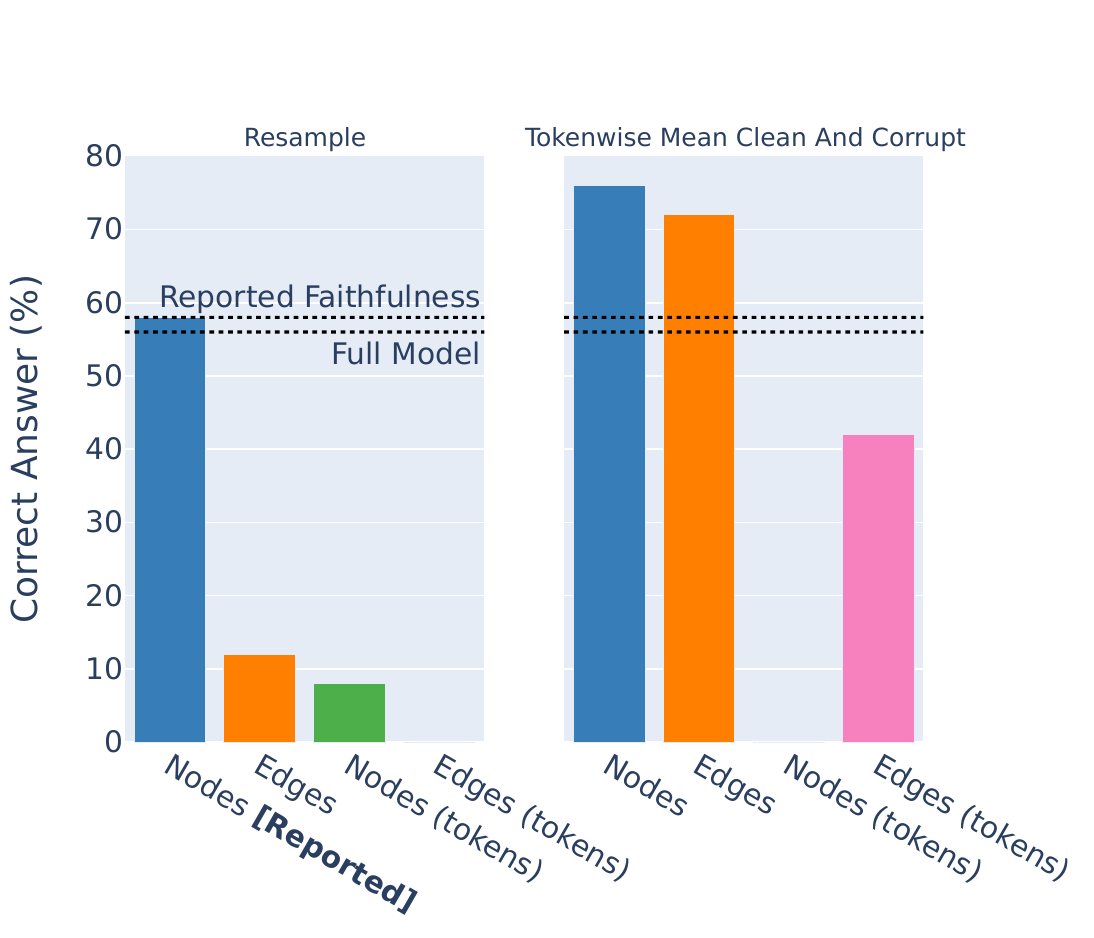}
        \caption{The faithfulness of the Docstring circuit according to the correct answer percent metric used by Heimersheim et al. \cite{docstring} is sensitive to the type of ablation used to measure the circuit.}
    \end{subfigure}
    \hfill
    \begin{subfigure}[b]{0.45\textwidth}
        \centering
        \includegraphics[width=\textwidth]{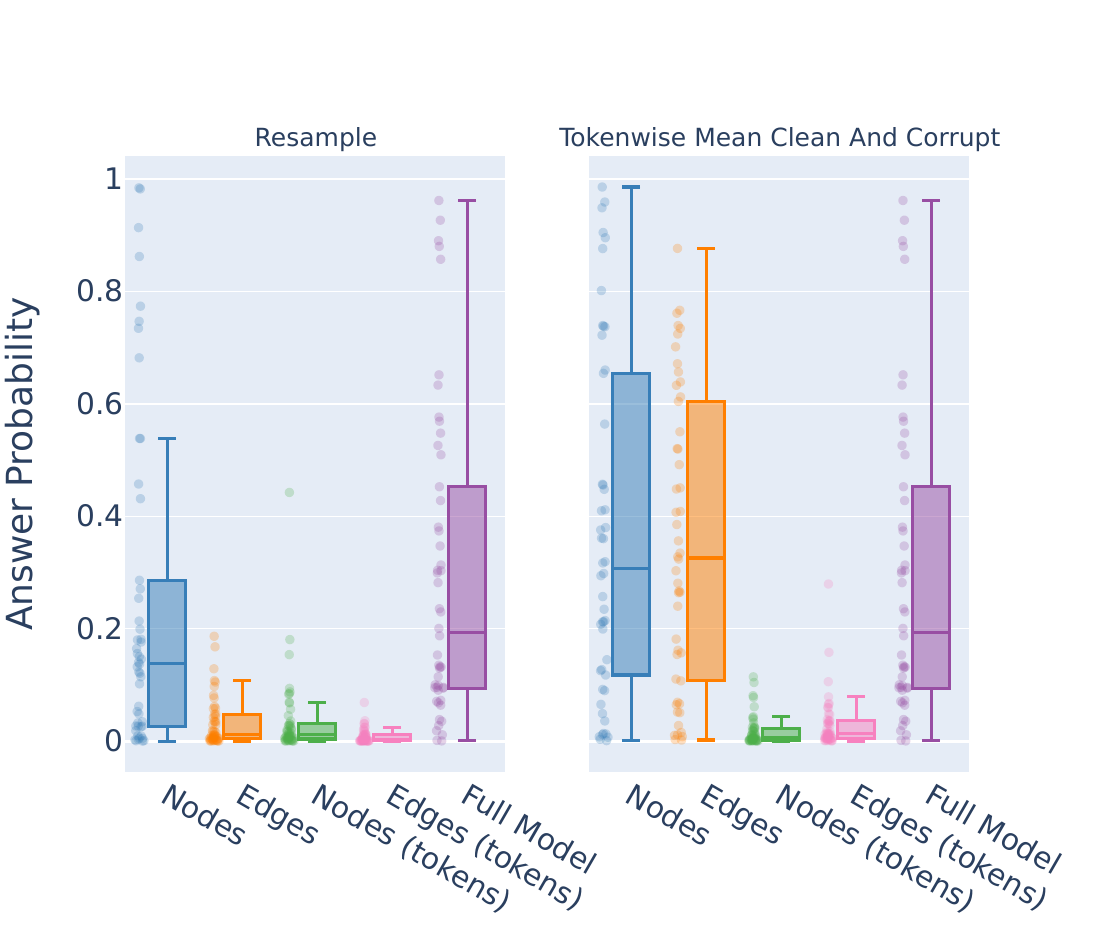}
        \caption{The faithfulness of the Docstring circuit as measured by the probability of the correct answer is highly variable between individual prompts.}
    \end{subfigure}
    \caption{Faithfulness metrics for the Docstring circuit when ablating every node or edge \textit{not} in the circuit, at all token positions and at token positions specified by \citet{docstring}.}
    \label{fig:docstring-faithfulness}
\end{figure}

In Figure~\ref{fig:docstring-faithfulness}, we test the faithfulness of the Docstring circuit with various ablation methodologies. We compare: (1) distinguishing between different token positions (Heimersheim \& Janiak specify their circuit with token positions, even though they do not use this information in their faithfulness evaluations), (2) ablating at the edge-level and node-level (they also specify edges, even though they evaluate only with nodes), (3) ablating with Resample and Mean Ablations and (4) two different faithfulness metrics: correct answer percentage and answer probability. 

We measure various significant changes in faithfulness in response to these adjustments. Most importantly, Edge Ablations perform significantly better using a Mean Ablation instead of a Resample Ablation. Had \citet{docstring} performed edge-level Resample Ablations instead of node-wise Resample Ablations, they may have trusted their circuit significantly less (and if they had used edge-level Mean Ablations, they may have trusted it more).

Distinguishing by token position also had a large effect on faithfulness scores for both node-wise and edge-wise ablations. These low scores suggest the circuit is in fact performing significant computation on token positions outside of the circuit specified by \citet{docstring}.

When we measure the probability of the correct answer we find that, similar to IOI, the variance between individual prompts is high. This is important for reasons outlined in Section~\ref{faithfulness}.

\subsection{Sports Players}

\begin{figure}[!ht]
    \centering
    \begin{subfigure}[b]{0.45\textwidth}
        \centering
        \includegraphics[width=\textwidth]{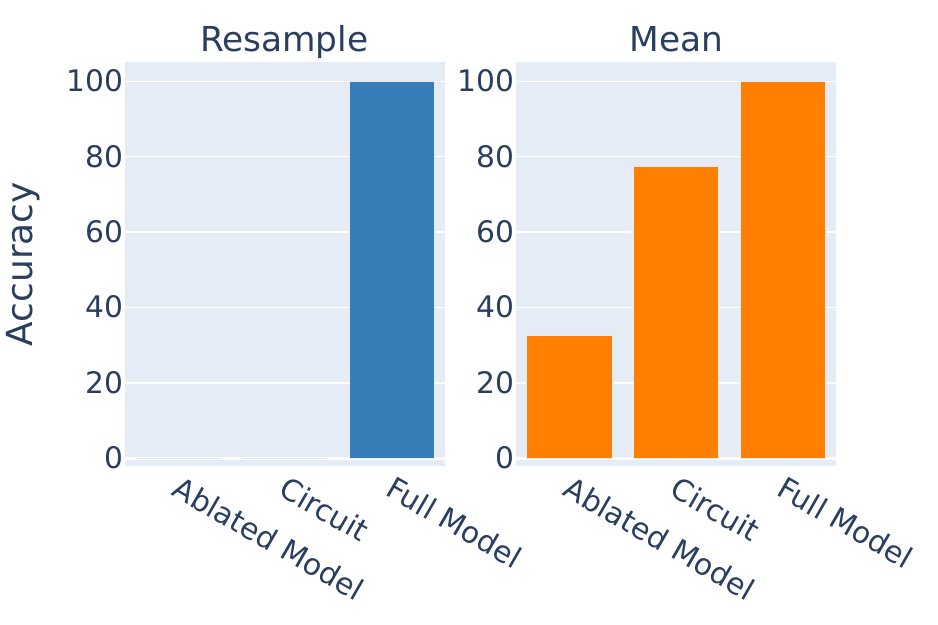}
        \caption{The percentage of prompts for which the correct sport has the highest output logit with Mean and Resample Ablations.}
    \end{subfigure}
    \hfill
    \begin{subfigure}[b]{0.45\textwidth}
        \centering
        \includegraphics[width=\textwidth]{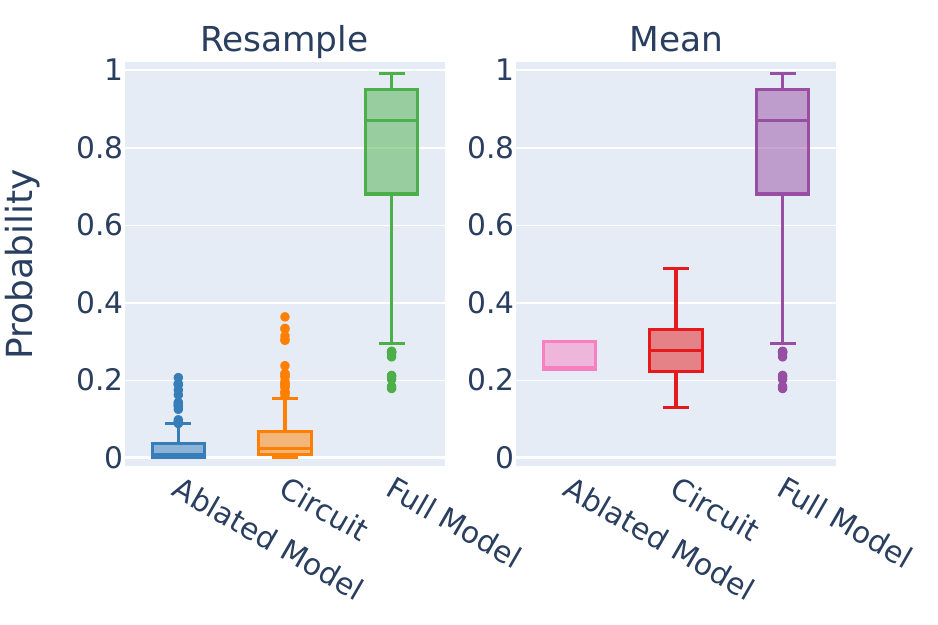}
        \caption{The output probability of the correct sport with Mean and Resample Ablations.}
    \end{subfigure}
    \caption{The faithfulness of the Sports Players circuit is reduced when using Resample Ablations.}
    \label{fig:sports-players-faithfulness}
\end{figure}

\textbf{The Sports Players Task}. The Sports Players task \citep{nanda2023factfinding} is a simple task that tests the Pythia-2.8b model's \citep{biderman2023pythia} ability to recall the sports of famous football, baseball and basketball players. See Table~\ref{tab:task-table} for an example. All prompts follow a very similar format, with the only difference being the name of the sports player in question. The corrupt distribution follows the exact same format, with each clean/corrupt pair having two players of different sports.

\textbf{Measuring Sports Players Circuit Faithfulness}.
In Figure~\ref{fig:sports-players-faithfulness}, we test the faithfulness of the edge-level sports players circuit, distinguishing token positions while (1) ablating the complement with both Resample and Mean Ablations and (2) calculating two different faithfulness metrics: correct answer percentage (considering only the three possible sports, following \citet{nanda2023factfinding}) and answer probability.

We find a dramatic difference in correct answer percentage between Resample and Mean Ablation. This case is a little different because the authors' aim wasn't to find the full circuit but to identify the place in the model where factual recall occurs, so this result doesn't negate their hypothesis.

Note that random guessing would achieve \(33\%\) accuracy as there are 3 possible sports, and this is roughly what we see when Mean Ablating the whole model. But Resample Ablating adds signal from the corrupt prompt, which is always a different sport, explaining the \(0\%\) accuracy score for the Ablated Model and the Circuit.

\clearpage
\subsection{Further Detail on the X-Proportion Tracr Ground Truth Circuits}

\begin{figure}[h]
    \centering
    \begin{subfigure}[b]{0.45\textwidth}
        \centering
        \includegraphics[width=\textwidth]{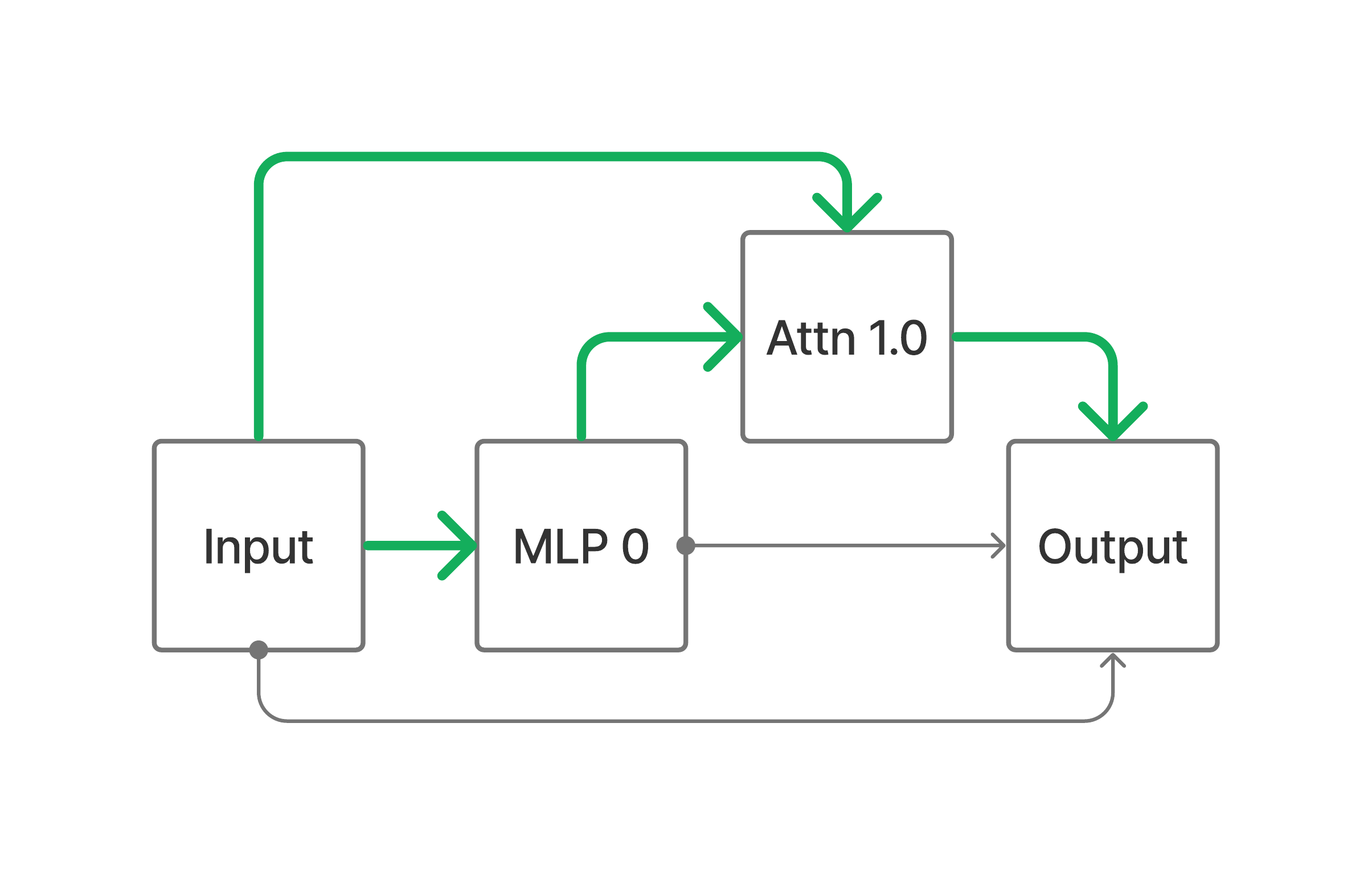}
        \caption*{(theirs) The ``ground truth'' circuit for the Tracr X-Proportion task using Zero Ablations.}
        \label{fig:xproportion-their-circ-diagram}
    \end{subfigure}
    \hfill
    \begin{subfigure}[b]{0.45\textwidth}
        \centering
        \includegraphics[width=\textwidth]{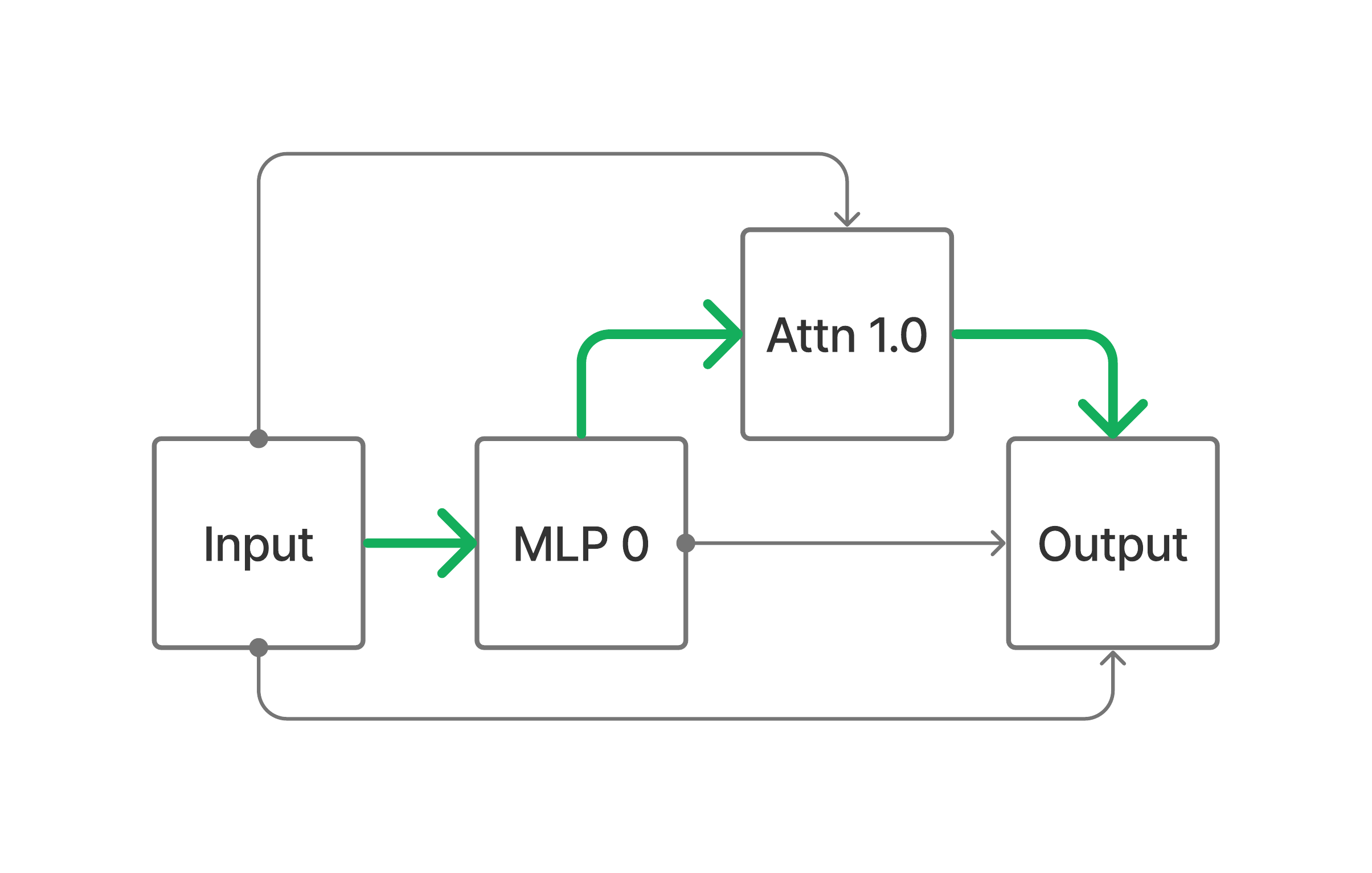}
        \caption*{(ours) The ``ground truth" circuit for the Tracr X-Proportion task using Resample Ablations.}
        \label{fig:xproportion-our-circ-diagram}
    \end{subfigure}
    \caption{For the Tracr X-Proportion circuit, the edge from \texttt{Input} to \texttt{Attn 1.0} is only used to transfer the positional encoding, so it is not required when using Resample Ablations, since these preserve information that is constant between the clean and corrupt distribution. This illustrates the principle that optimal circuits cannot be defined without an ablation methodology. (Nodes \texttt{Attn 0.0} and \texttt{MLP 1} are not shown as they are not used in this model.)}
    \label{fig:xproportion-diagrams}
\end{figure}

\clearpage
\section{Edge-Based vs. Node-Based Circuit Discovery Methods} \label{edge-vs-node-circuit-discovery}

\begin{figure}[!ht]
    \centering
    \includegraphics[width=0.7\textwidth]{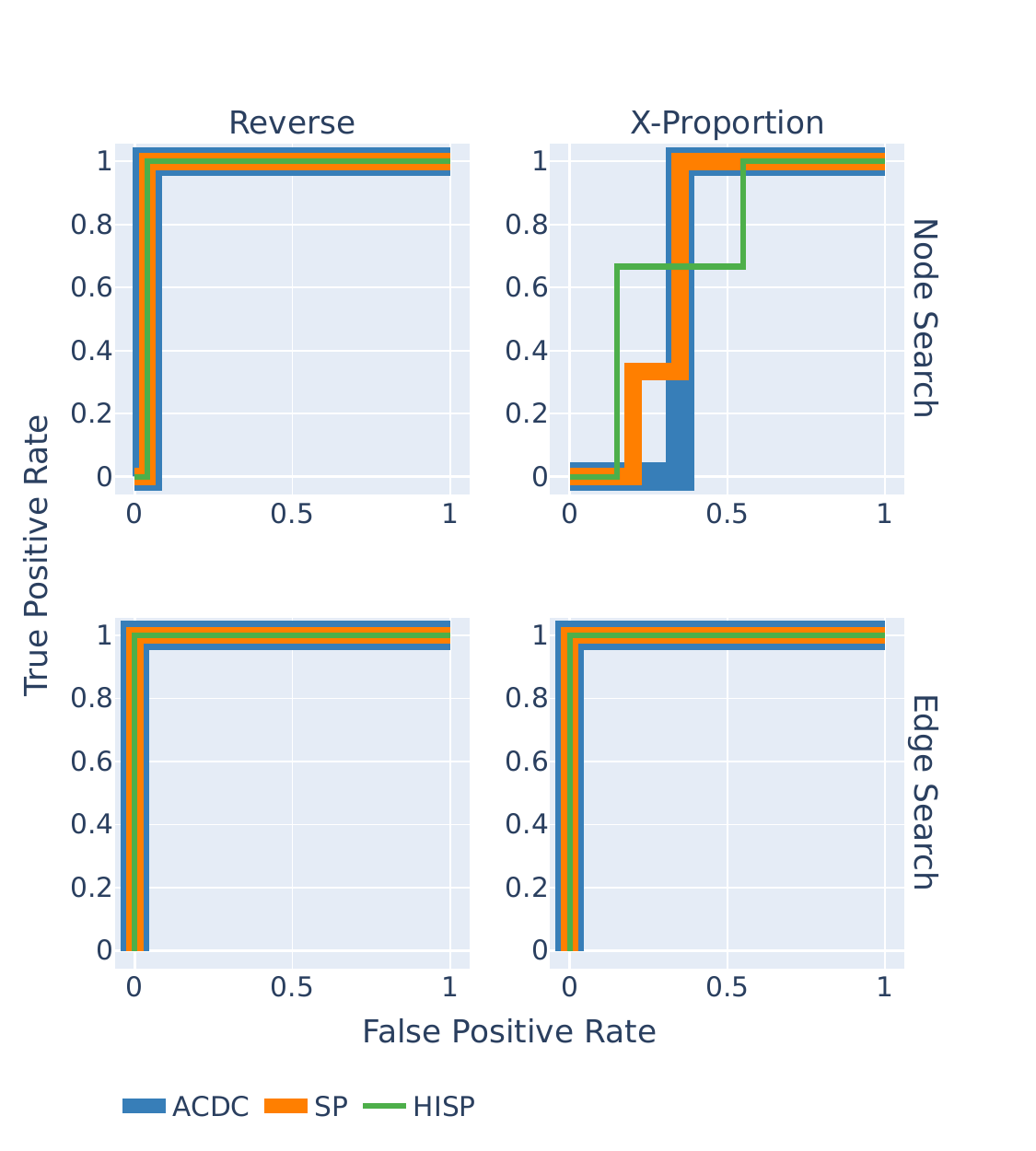}
    \caption{ROC Curves for Edge-Based and Node-Based circuit discovery methods, using the Resample Ablation edges as the ground truth (ours).}
    \label{fig:edge-vs-node-circuit-discovery}
\end{figure}

In Section~\ref{optimal-circuits}, we adapted the Subnetwork Probing (SP) and Head Importance Scoring (HISP) circuit discovery methods to use (or approximate) Edge Ablation. ACDC \citep{conmy2023automated} already uses Edge Ablations, but we can similarly adapt ACDC to use Node Ablations. We compare the performance of the Node Patching versions of ACDC, SP and HISP to the Edge Patching versions, for the Resample Ablation based ``ground truth" circuit introduced in Section~\ref{optimal-circuits} (Figure~\ref{edge-vs-node-circuit-discovery}).

\clearpage
\section{Clarifying Nomenclature}

Some authors have used different terms for some of the concepts introduced in Section~\ref{ablation-in-language-models}. For instance, Activation patching has previously also been called Causal Tracing or Interchange Intervention. In the remainder of this section, we summarise how our nomenclature relates to the terminology used by Redwood Research in their series of early mechanistic interpretability transformer-circuits papers. Chronologically, these are \citet{wang2022interpretability, causal_scrubbing, nix_path_patching}.

We first discuss the final, most comprehensive work \citep{causal_scrubbing}, which we refer to as Causal Scrubbing. Causal Scrubbing is a very general approach for evaluating circuits together with explanations of the role of nodes within the circuit. It generically comprises performing specific branch-based Resample Ablations on the treeified model on both the circuit \textit{and} its complement. Causal Scrubbing randomly replaces activations with those that your hypothesis predicts will not change the model output. For instance, if we claim that a given node detects whether the input is even, Causal Scrubbing could patch in an activation from a different even input, and expects the output not to change. In general, Causal Scrubbing permits an arbitrary number of possible counterfactual inputs.

\citet{nix_path_patching} simplify this setup, dropping the strict requirement of requiring an explanation for each node. This reduces the hypothesis class to the now standard circuit discovery problem; does some path matter for task performance or not?

Finally \citet{wang2022interpretability} perform a further simplified version of path patching to discover the IOI circuit. This is equivalent to Edge Resample Ablation in our terminology but which they call Path Patching. They patch paths one at a time, to establish which edges are important for task performance. Importantly, \citet{wang2022interpretability} reason that the IOI task should be an attention-only task, as it only comprises moving information between tokens. As such, they take nodes to only be attention heads, with MLPs considered to be part of the direct path between nodes. This approach of one-hop path patching is extended and automated by \citet{conmy2023automated}.

\end{document}